%% file: digipro.tex
\documentclass[sigconf]{acmart} 
\renewcommand\footnotetextcopyrightpermission[1]{}
\usepackage{amsmath}

\DeclareMathOperator*{\argmin}{arg\,min}

\copyrightyear{2022}
\acmYear{2022}
\setcopyright{acmcopyright}\acmConference[DigiPro '22]{The Digital Production Symposium}{August 7, 2022}{Vancouver, BC, Canada}
\acmBooktitle{The Digital Production Symposium (DigiPro '22), August 7, 2022, Vancouver, BC, Canada}
\acmPrice{15.00}
\acmDOI{10.1145/3543664.3543681}
\acmISBN{978-1-4503-9418-5/22/08}

\begin{document}

\title{Jointly Optimizing Color Rendition and In-Camera Backgrounds in an RGB Virtual Production Stage}

\author{Chloe LeGendre}
\affiliation{
  \institution{Netflix}
  \country{USA}
}
\email{clegendre@netflix.com}

\author{Lukas Lepicovsky}
\affiliation{
  \institution{Scanline VFX}
  \country{USA}
}
\email{lukas.lepicovsky@scanlinevfx.com}

\author{Paul Debevec}
\affiliation{
  \institution{Netflix}
  \country{USA}
}
\email{debevec@netflix.com}

\renewcommand{\shortauthors}{LeGendre, Lepicovsky, and Debevec}

\begin{abstract}

While the LED panels used in today's virtual production systems can display vibrant imagery within a wide color gamut, they produce problematic color shifts when used as lighting due to their "peaky" spectral output from narrow-band red, green, and blue LEDs.  In this work, we present an improved color calibration process for virtual production stages which ameliorates this color rendition problem while also maintaining accurate in-camera background colors.  We do this by optimizing linear color correction transformations for 1) the LED panel pixels visible in the camera's field of view, 2) the pixels outside the camera's field of view illuminating the subjects, and – as a post-process – 3) the pixel values recorded by the studio camera. The result is that footage shot in an RGB LED panel virtual production stage can exhibit more accurate skin tones and costume colors while still reproducing the desired colors of the in-camera background.


\end{abstract}



\begin{teaserfigure}
  \vspace{-6pt}
  \begin{center}
  \begin{tabular}{c@{ }c@{ }c}
    \includegraphics[width=2.25in]{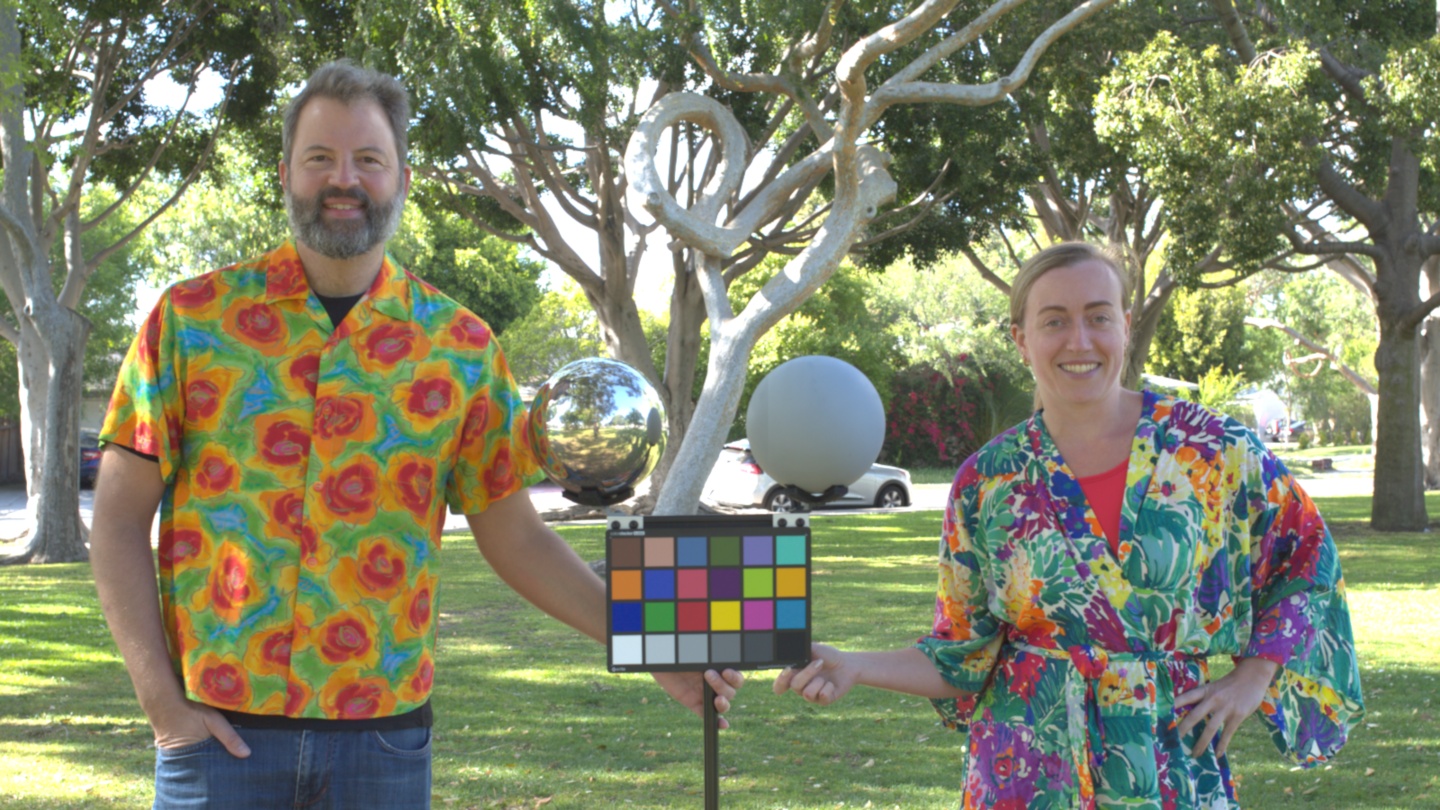} & 
    \includegraphics[width=2.25in]{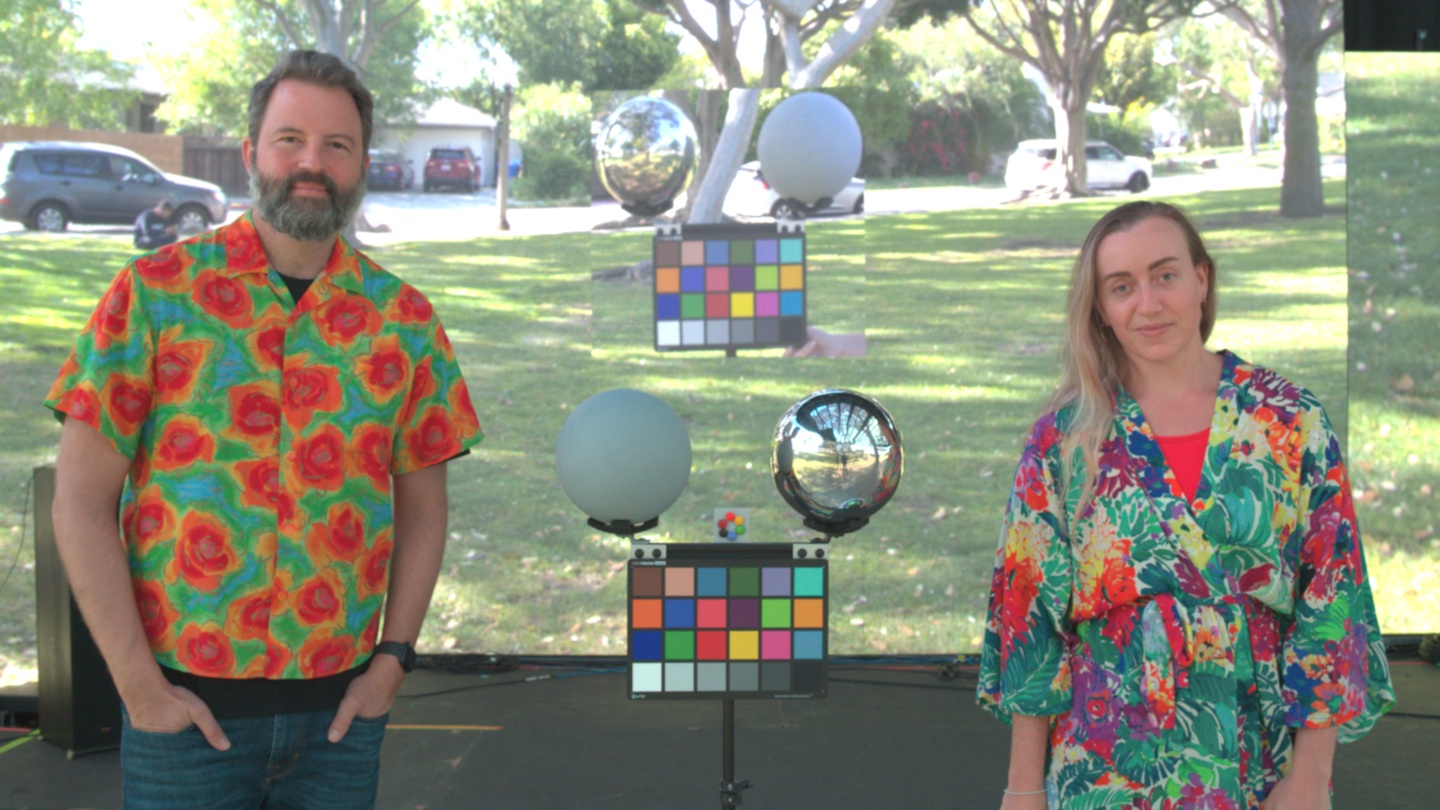} &
    \includegraphics[width=2.25in]{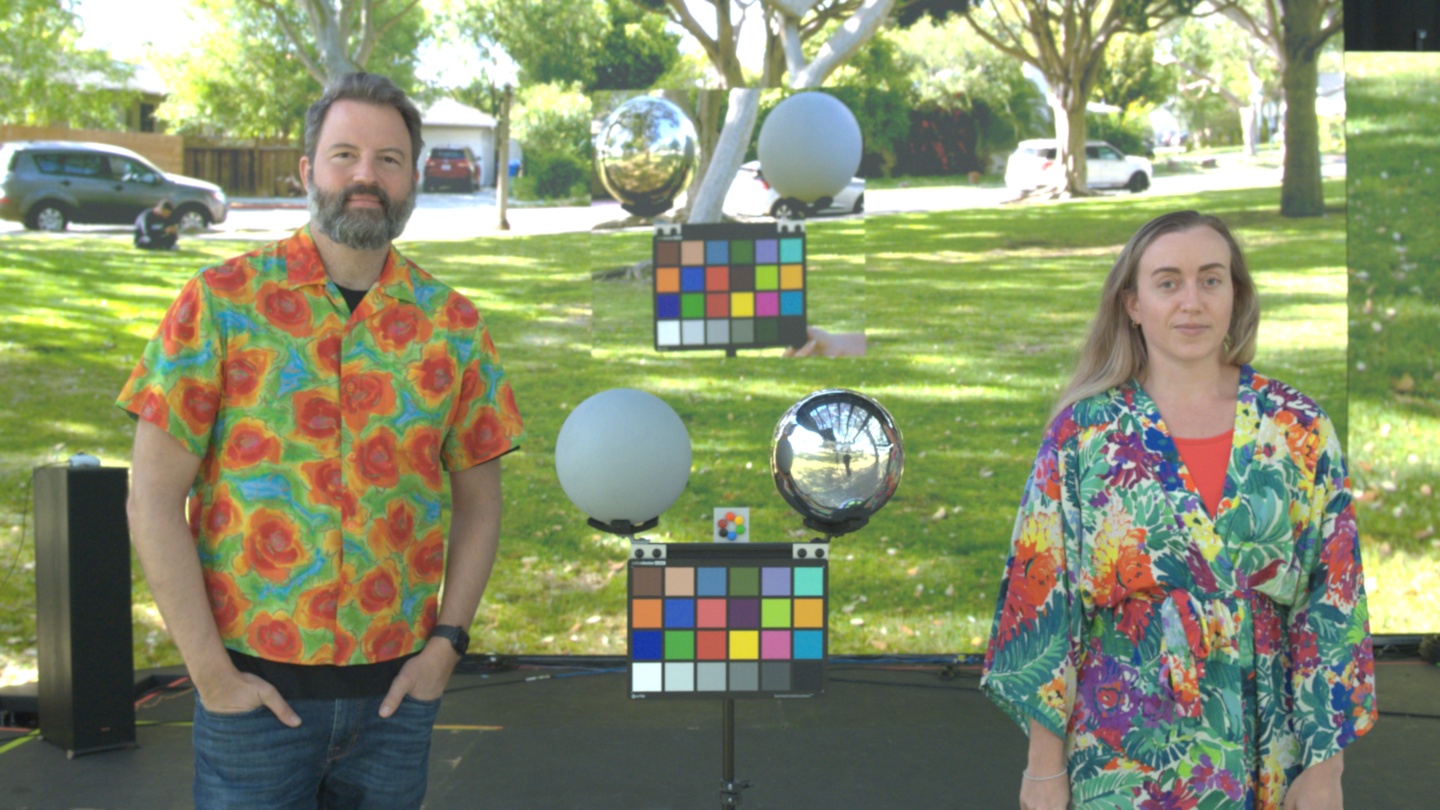} \\
    \footnotesize{(a) subjects lit by real outdoor environment} &
    \footnotesize{(b) subjects lit by calibrated VP stage (baseline)} &
    \footnotesize{(c) subjects lit by calibrated VP stage (our method)} \\
    \end{tabular}
    \end{center}
  \vspace{-7pt}
  \caption{(a) Subjects in colorful clothing with a color chart and lighting reference spheres in an outdoor environment. (b) The same subjects in an RGB LED virtual production stage lit by color-matched imagery of the environment, showing color rendition errors in clothing colors and skin tones. (c) The subjects in an RGB LED virtual production stage calibrated using our technique to optimize both color rendition accuracy and in-camera background appearance.}
  \vspace{3pt}
  \label{fig:teaser}
\end{teaserfigure}


\maketitle
\pagestyle{plain}

\section{Introduction and Related Work}

\subsubsection*{Virtual Production}
Although the term \textit{Virtual Production} (VP) refers to a cadre of novel, technology-driven film-making methods, many recent incarnations employ LED stages or "volumes" to surround actors with imagery of virtual film locations. This technique, cast as an alternative to filming in a greenscreen studio or on location, has exploded in popularity for both film and television production \cite{Kadner:2021b, Kadner:2021c, Holben:2020}. One benefit of the technique is that an actor may be photographed against a high-resolution background image shown on the LED panels, called filming an \textit{in-camera background}. This process can remove significant post-production compositing work as compared with the traditional workflows involving chromakeying, rotoscoping, and matting.

Beyond the in-camera background, an additional goal of filming inside an LED volume is to perform \textit{lighting reproduction}, where individual light sources surrounding an actor are driven to reproduce the illumination of a given scene \cite{Debevec:2002, Hamon:2014}. When displaying a high dynamic range, image-based lighting (HDR IBL) environment \cite{Debevec:1998}, a lighting reproduction system can generally match the subject's appearance to how they would appear in a real-world environment. An added benefit is that actors may feel more immersed in the scene, potentially enhancing their performances and providing natural eyelines compared to filming in a greenscreen studio.

\subsubsection*{Color Rendition Challenges}
\begin{figure}[h]
\vspace{-5pt}
\begin{tabular}{c@{ }c}
\includegraphics[width=1.3in]{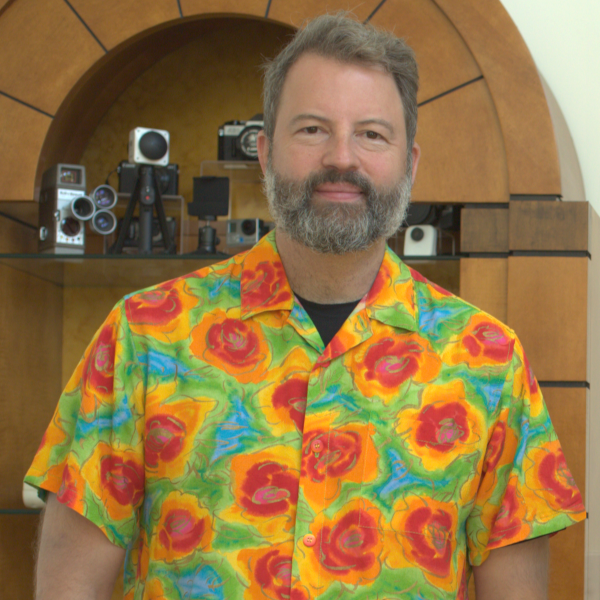} & 
\includegraphics[width=1.3in]{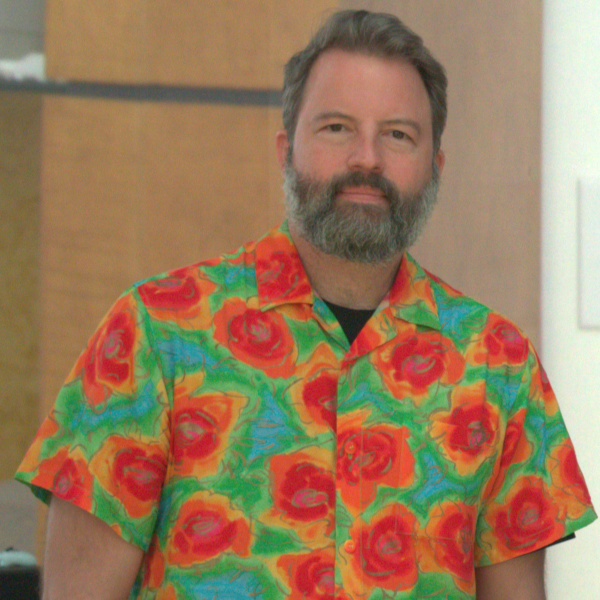} \\
\small{(a) indoors, natural light} & \small{(b) RGB LED VP stage} \\
\end{tabular}
\vspace{-5pt}
\caption{(a) a subject illuminated by indoor daytime natural lighting; (b) reproducing the illumination inside an RGB LED VP stage, with noticeable color rendition errors for skin tones and yellow and orange patches on the subject's shirt.}
\vspace{-7pt}
\label{fig:rgb_vp_example}
\end{figure}

Although LED volumes have enabled a whole new way to produce filmed content, cinematographers are becoming increasingly aware of color rendition challenges that come with using these systems for lighting reproduction, as recently noted and observed by Noah Kadner and Craig Kief in their \textit{American Cinematographer} article "Color Fidelity in LED Volumes" \cite{Kadner:2021a}. Most notably, in an LED volume, lighter skin tones shift towards pink and darker skin tones towards red. Orange materials also shift toward red, cyan materials shift toward blue, and yellow materials darken. For an example, see Fig. \ref{fig:rgb_vp_example}. 

\subsubsection*{The Unusual Spectra of RGB LEDs}
\begin{figure}[h]
\vspace{0pt}
\begin{tabular}{c@{ }c}
\small{(a) daylight (D65)} & \small{(b) cool white LED} \\
\includegraphics[height=1.05in]{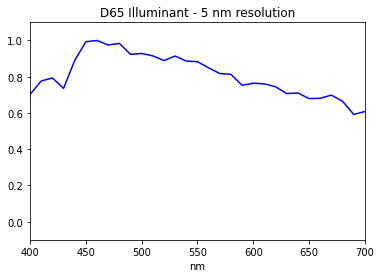} &
\includegraphics[height=1.05in]{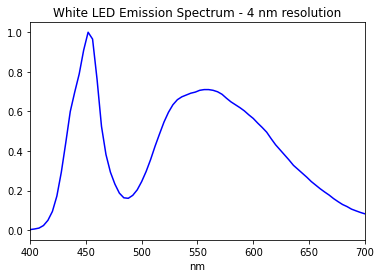} \\
\includegraphics[height=1.05in]{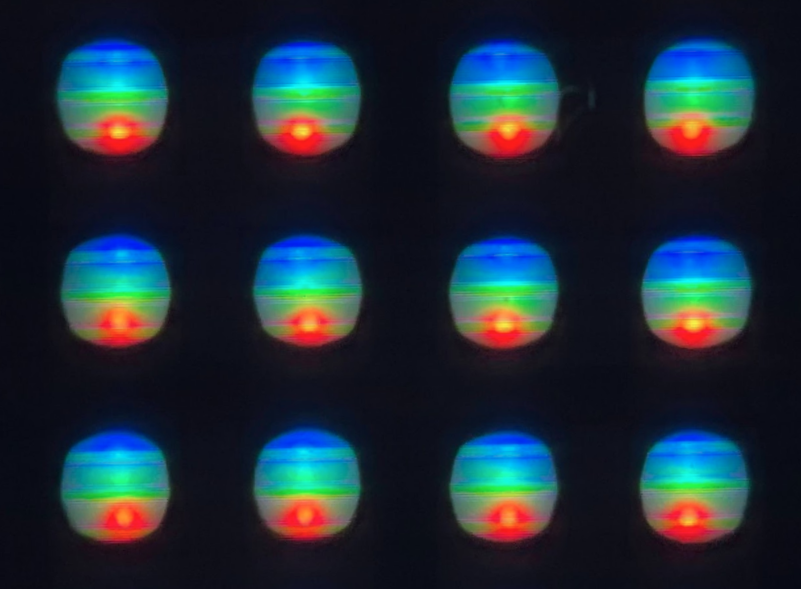} & 
\includegraphics[height=1.05in]{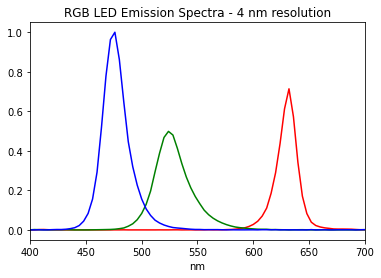} \\
\small{(c) RGB LED units} & \small{(d) RGB LED} \\
\end{tabular}
\vspace{-7pt}
\caption{The emission spectra of (a) daylight (D65); (b) a broad-spectrum cool white LED; and (d) a typical RGB LED, with a zoomed-in view of an LED panel showing an RGB LED package at each pixel in (c).}
\vspace{-10pt}
\label{fig:illuminants}
\end{figure}

These color rendition errors are the result of the unusual emission spectra of the RGB LEDs that comprise LED panel based VP stages. Unfortunately, while RGB LED panels have been designed for and perform well at displaying imagery within a wide gamut of \textit{colors}, they cannot reproduce any desired target illuminant \textit{spectrum}. In particular, the emission spectra of most real-world illuminants (e.g. daylight and white LEDs, see Fig. \ref{fig:illuminants}) are relatively broad, covering most of the visible light wavelength range. In contrast, by combining different amounts of light produced by RGB LEDs, one will only be able to produce illumination with relatively ``peaky" emission spectra, with distinct gaps between the spectra of each LED channel [see Fig. \ref{fig:illuminants}(d)]. Because the world acts as a "spectral renderer," color rendition errors for yellow and cyan materials are the result of these gaps in the LED emission spectra, while for skin tones an added culprit is the relatively long wavelength of red LEDs that illuminate skin where it is considerably more reflective. 

\subsubsection*{Filling in Spectral Gaps}
For a single light source and an omnidirectional light stage respectively, Wenger et al. \shortcite{Wenger:2003} and LeGendre et al. \shortcite{LeGendre:2016, LeGendre:2017} demonstrated improved color rendition by adding spectral channels beyond RGB to fill in these spectral gaps. Similarly, cinema light source manufacturers also incorporate broad-spectrum LEDs (e.g. white) in their RGB-based light sources to improve color rendition. Unfortunately, these are essentially no manufactured LED panel manufacturers that have followed suit, as panel designs are optimized for a wide display gamut, where adding a broad-spectrum white LED is unnecessary as its color can already be produced by mixing RGB LEDs.

\subsubsection*{LED Panel Calibration Workflows}
Without adding spectral channels, digital imaging technologists color calibrating a typical RGB LED VP stage aim to make colors in the content to be displayed actually appear those colors to the camera. This calibration process considers both the spectral sensitivity of the motion picture camera and the emission spectra of the RGB LEDs, although it typically requires no spectral measurements. In similar processes described by Unreal Engine \shortcite{unreal:2022} and by Weta Digital (sec. 3.3.3.1, Weidlich et al. \shortcite{Weidlich:2021}), a patch of each color primary displayed by the panels is recorded by the camera. Then, a color transform (often a $3\times3$ matrix) is computed to apply to the input RGB pixel values in the content to be displayed, to ensure that the camera-recorded color primaries match the content's color primaries. As noted in Debevec et al. \shortcite{Debevec:2002}, if the primaries match, then all other colors including white must match as well, as all others are simply a linear combination of the three primaries.

This primary-based calibration process is ideal when the goal is simply to use the LED panels as a display system, say, for filming an in-camera background. However, difficulties arise as these panels are used for lighting reproduction. Now we must consider how they \textit{illuminate} actors and their costumes. 

\subsubsection*{Using a Calibrated RGB LED Panel as a Light Source}
\begin{figure}[h]
\vspace{0pt}
\begin{tabular}{c@{ }c@{ }c}
\includegraphics[width=1.in]{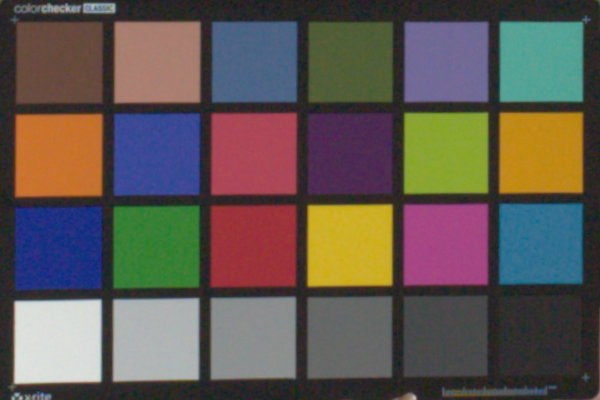} & 
\includegraphics[width=1.in]{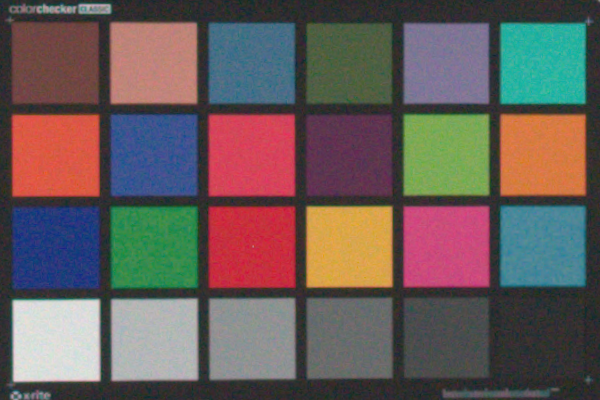} & 
\includegraphics[width=1.in]{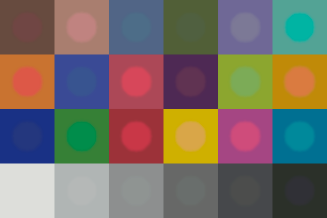} \\
\small{(a) real environment} & \small{(b) RGB LED VP} & \small{(c) sampled from (a, b)}\\
\end{tabular}
\vspace{-7pt}
\caption{(a) a color chart photographed in a real outdoor environment; (b) photographed in a VP stage reproducing the target illumination using RGB LEDs and a standard primary-based calibration process; (c) comparison between (a) and (b). Background squares are pixel values sampled from (a) and foreground circles are pixel values sampled from (b). Although the white squares match (the illuminants are \textit{metameric matches}), the remaining color squares do not.}
\vspace{-5pt}
\label{fig:wb_example}
\end{figure}

As an example, say we photograph a color chart lit by daylight (see Fig. \ref{fig:wb_example}a) along with the scene's corresponding HDR IBL. From the chart's white square, we can extract an RGB pixel value $w$ that represents the average color of the scene's illumination (specifically, for the hemisphere of lighting directions facing the color chart). We know that a color chart placed in a VP stage displaying the HDR IBL, calibrated using the above technique, will have a white square value that matches $w$ (see Fig. \ref{fig:wb_example}b). In this case, the real and reproduced illuminants are \textit{metamers}: matching colors with non-matching spectral power distributions. However, the \textit{spectrum} of the lighting reproduced in the VP stage will be quite different from that of real-world daylight, and so the colors of any non spectrally flat / non-neutral color chart squares will be unlikely to match. The comparison visualization in Fig. \ref{fig:wb_example}c shows this effect. A virtual color chart \textit{displayed} by the panels would appear largely correct; only the real color chart \textit{lit} by the LED panels would exhibit these color rendition errors. 

\subsubsection*{Our Approach}
With these color rendition challenges in mind, we propose a novel technique for the color calibration of an RGB LED stage using a \textit{set} of linear transformations represented as $3\times3$ color correction matrices. Our primary goal is to optimize the color rendition properties of the LED volume acting as a \textit{light source}, while still maintaining accurate in-camera background colors. Since we know that RGB LEDs produce color rendition errors when simulating broad-spectrum lighting, we solve for an optimal post-correction matrix to improve color matching. Such a post-correction matrix can desaturate overly pink/red skin tones while keeping neutral colors neutral, which cannot be accomplished by altering the content displayed on the LED stage alone. We further show that if we apply the inverse of the post-correction matrix to the in-camera-frustum area, then the in-camera background color matching can be maintained as well. While prior calibration approaches focus exclusively on in-camera background color accuracy and do not consider post-correction strategies, our multi-matrix approach ensures near optimal color rendition simultaneously both for foreground content (e.g. actors, set, costumes) as well as the in-camera background. Our technique requires no spectral measurements of any part of the system (camera, materials, or LED panels), relying only on four calibration images captured with the principal-photography camera for a given LED volume.

\section{Method}

\subsection{Overview of Our Approach}
\begin{figure}[ht]
\vspace{0pt}
\begin{tabular}{c}
\includegraphics[width=3.3in]{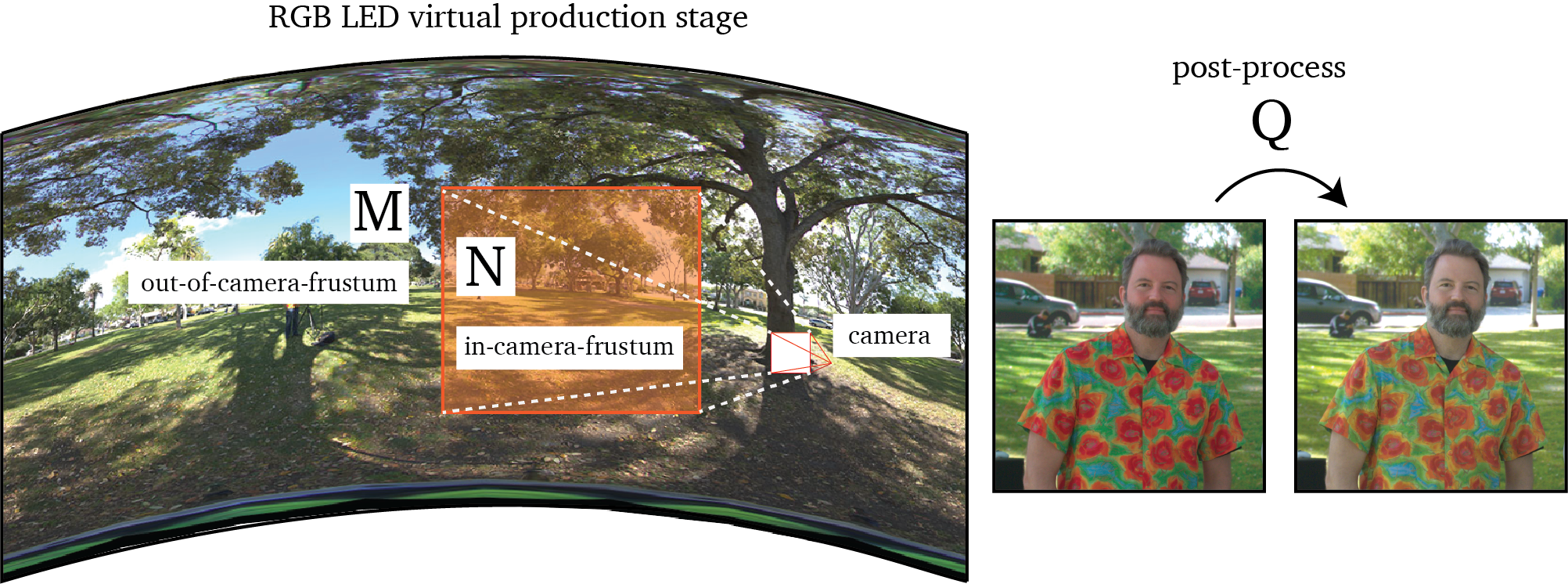}  \\
\end{tabular}
\vspace{-5pt}
\caption{A visualization of our method. A $3 \times 3$ pre-correction matrix $\textbf{M}$ is applied for out-of-frustum content, while a different $3 \times 3$ pre-correction matrix $\textbf{N}= \textbf{M}{\textbf{Q}}^{-1}$ is applied for in-camera-frustum content. Finally, a $3 \times 3$ post-correction matrix $\textbf{Q}$ is applied to the recorded image content.}
\vspace{-5pt}
\label{fig:methodology}
\end{figure}

To summarize, we propose the following straightforward steps:
\begin{enumerate}
    \item We solve for a $3\times3$ \textit{pre-correction} matrix $\textbf{M}$ (to be applied to the displayed content) that maps the target scene's pixel colors to the LED panel colors, so they look the same to the motion picture camera. In general, applying $\textbf{M}$ to the displayed content will not, however, \textit{light} the stage subjects accurately, as colors will appear overly-saturated.
    \item We thus solve for $3\times3$ \textit{post-correction} matrix $\textbf{Q}$ (to be applied to the final recorded image) that will make a photographed color chart lit by the VP stage displaying the HDR IBL environment look as close as possible to how it appeared in the real scene. However, if we apply $\textbf{Q}$ to the whole image, the in-camera background pixels will no longer appear correct.
    \item Thus, instead of applying $\textbf{M}$ to the in-camera-frustum content, we apply a \textit{different} $3\times3$ pre-correction matrix $\textbf{N}= \textbf{M}{\textbf{Q}}^{-1}$. As the background pixels do not contribute significantly to the lighting on the actors, both the color rendition on the actors and the appearance of the in-camera pixels will be near optimal. To visualize our methodology, see Fig. \ref{fig:methodology}. 
\end{enumerate}

\subsection{Assumptions and Prerequisites}

\subsubsection*{Panel and Camera Linearity}
We first assume that an LED volume has been calibrated to act as a linear display. Images of displayed color swatch ramps, or an image series shot with increasing pixel values on the LEDs \cite{Debevec:2002} can be used to verify linearity, and if needed, to correct for it. Our method also assumes that the camera used throughout the imaging workflow also has a linear response, as is typical for digital cinema cameras. In this work, we verified the LED panels showed images with a gamma value of 2.4, and gamma-corrected all linearly computed images for display accordingly.

\subsubsection*{Radiometric Alignment of Different Panel Types}
LED volumes in practice are often comprised of multiple types of LED panels, depending on whether they are designed to cover a studio's ceiling, wall, or floor. Our method also further assumes that the relative brightness levels of different panel types comprising an LED volume have been calibrated such that a pixel value of $[1, 1, 1]$ displayed from all directions produces as uniform as possible of a sphere of light of even intensity and color balance from all directions.

\subsubsection*{HDRI Map Acquisition and Display}
We assume that the lighting environment to be displayed on the VP stage will be captured using HDR panoramic photography techniques (e.g. \cite{Debevec:1998}). We further assume that a color chart has been photographed at a spatial location such that the HDR panorama's center of projection matches the location of the color chart in the scene. In practice, this means that our technique could be less accurate for color charts shot at some distance from the HDRI map's location. Finally, we assume that the VP stage is capable of representing the full dynamic range of the HDRI map, without clipping any light sources.

\subsection{Calibration Images and Equations}
Our technique requires just four calibration images, all photographed using the target camera to be used for filming in the LED volume. We next describe these photographs, providing intuition as to why they are needed. The first is of the LED panels displaying color patches for each primary color, required for computing $\textbf{M}$. The second, third, and fourth images record how each spectral channel of the LED stage \textit{lights} a color chart, required for computing $\textbf{Q}$.

\subsubsection*{Solving for $\textbf{M}$: Primary-based Calibration}

\begin{figure}[ht]
\vspace{-5pt}
\begin{tabular}{c@{ }c@{ }c}
\includegraphics[width=1.in]{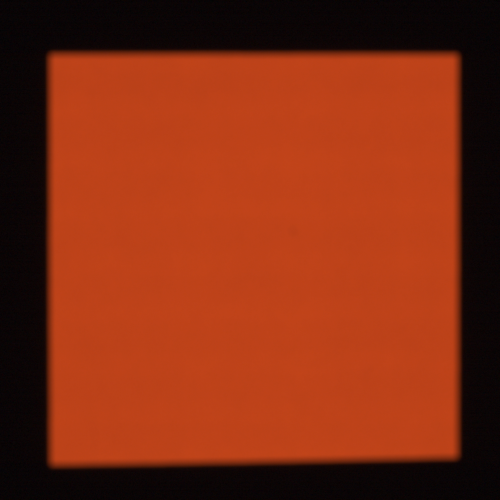} & 
\includegraphics[width=1.in]{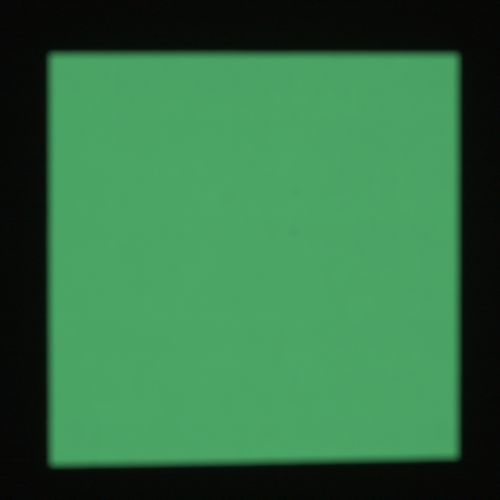} & 
\includegraphics[width=1.in]{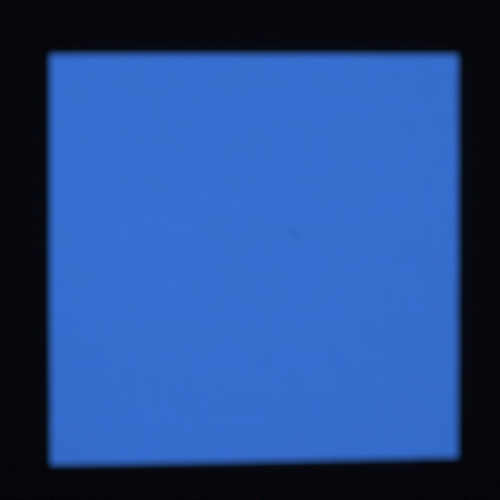} \\
\end{tabular}
\vspace{-5pt}
\caption{The red, green, and blue LEDs as observed by our camera. Pixel values sampled from these images form matrix $\textbf{[SL]}$, encoding how the camera observes the LED primaries.}
\vspace{-7pt}
\label{fig:mim_calib}
\end{figure}

This measurement allows us to map the target scene's pixel colors to LED panel colors observed by the camera, ensuring a metameric illuminant match when a scene's HDRI map is displayed. For this calibration, we display a patch of pure red, pure green, and pure blue on the in-camera-frustum LED panels and record their appearance to the camera (see Fig. \ref{fig:mim_calib}). These patches can all be photographed in a single image.

This is the same process that is used to generate a primary-calibrated LED panel \cite{Weidlich:2021, unreal:2022}. From these images, we extract average pixel values and concatenate them along columns to obtain a $3\times3$ matrix that we call $\textbf{[SL]}$, because its elements are the pairwise dot products of the camera's \textbf{S}pectral sensitivity functions and the \textbf{L}ED emission spectra. $\textbf{[SL]}$ has the camera's color channels along rows, and the spectral channels of the LED volume along columns. We ensure that out-of-frustum panels are turned off during this capture, to prevent including light bounced off the front side of the panels in the measurement. From $\textbf{[SL]}$, we can solve for our $3\times3$ matrix $\textbf{M}$:

\vspace{-5pt}
\begin{equation}
    \textbf{[SL]MI = I}
    \label{eqn:m_solve0}
\end{equation}

Here, $\textbf{I}$ is the identity matrix. This equation holds because our goal is that pixel values corresponding to pure red, green, and blue ($[1, 0, 0], [0, 1, 0]$ and $[0, 0, 1]$) displayed by the panel are observed as the same pixel values to the camera. We can also think of this equation as linearly combining the LED primaries as seen by the camera (columns of $\textbf{[SL]}$) to produce the final pixel values of our image. We can easily solve for $\textbf{M}$ using matrix inversion:

\vspace{-3pt}
\begin{equation}
    {\textbf{M = [SL]}}^{-1}
\end{equation}

Thus far, this process is nearly identical to that used for calibrating a typical VP stage for in-camera VFX. However, in our technique we will use $\textbf{M}$ as a color correction matrix for out-of-frustum content only, rather than for all content as before.

\subsubsection*{Solving for post-correcting $\textbf{Q}$: Color Rendition Calibration}

Next, our goal is to solve for a $3\times3$ \textit{post-correction} matrix $\textbf{Q}$ which, when applied to the final image, makes a color chart lit by the VP stage displaying the HDR IBL environment look as close as possible to how it appeared in the real-world scene. It is common practice to capture a slate including a color chart during filming, so we assume that such imagery will be available. While we could, in practice, just photograph the chart as illuminated by the HDR IBL displayed by the panels (pre-corrected with $\textbf{M}$), and solve for $\textbf{Q}$ from this image directly, ideally we would be able to compute $\textbf{Q}$ only from calibration imagery, captured once regardless of the number of lighting environments that we wish to display. 

Towards this end, we photograph the color chart as illuminated by each spectral channel of the LED volume individually, following the procedure outlined by LeGendre et al. \shortcite{LeGendre:2016}. The core insight here is that because of the superposition principle for light, any chart illuminated by the VP stage will resemble a linear combination of these three images. So, if we capture such data, we can \textit{simulate} the appearance of a color chart illuminated by any given environment, rather than needing to photograph it each time to compute $\textbf{Q}$. 

\begin{figure}[ht]
\vspace{-5pt}
\begin{tabular}{c@{ }c@{ }c@{ }c}
{\rotatebox{90}{\footnotesize{calibration setup}}} &
\includegraphics[width=1.in]{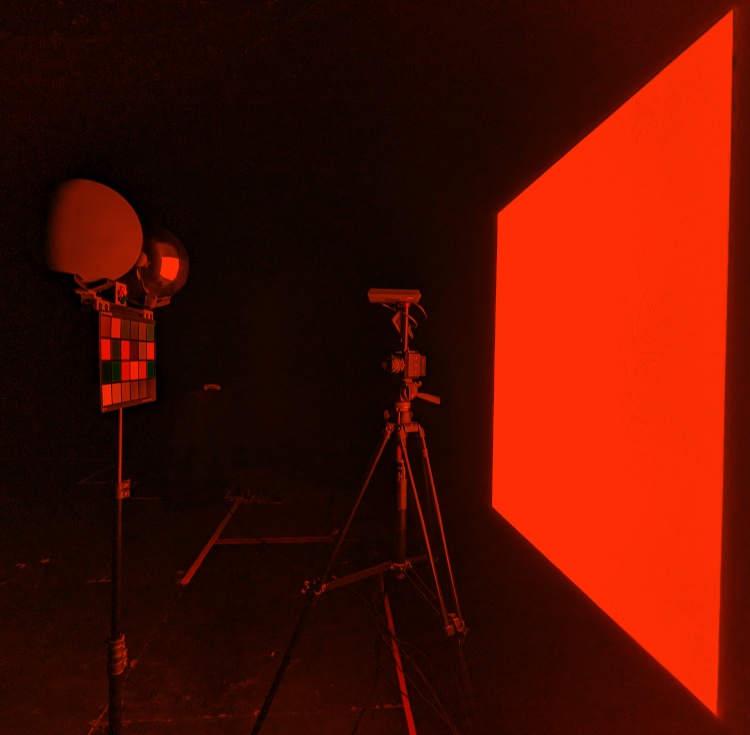} & 
\includegraphics[width=1.in]{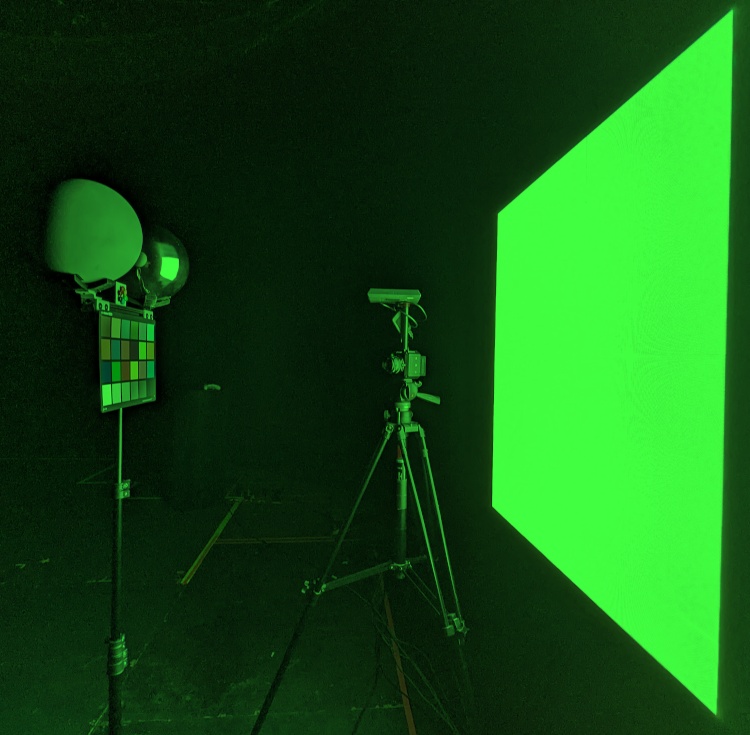} & 
\includegraphics[width=1.in]{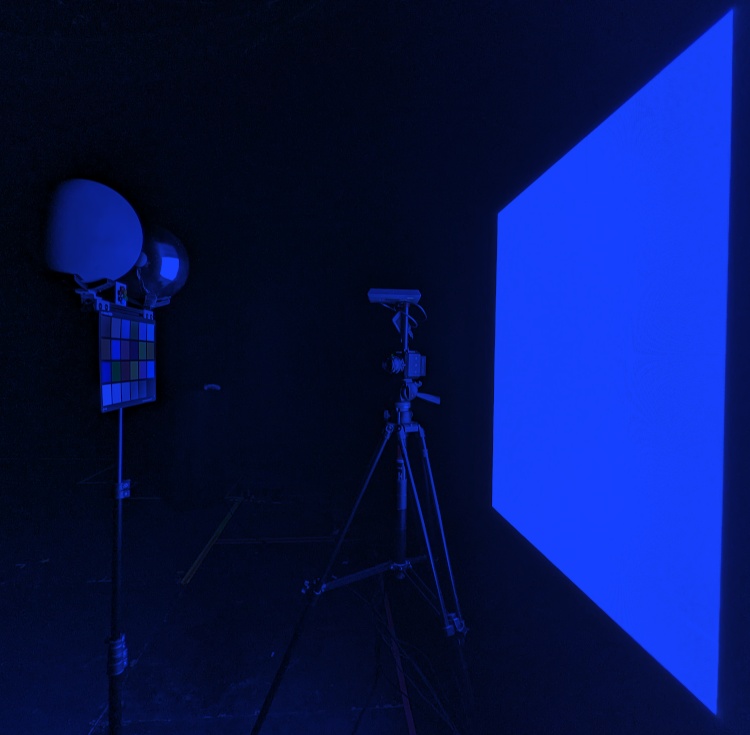} \\
{\rotatebox{90}{\footnotesize{photograph}}} &
\includegraphics[width=1.in]{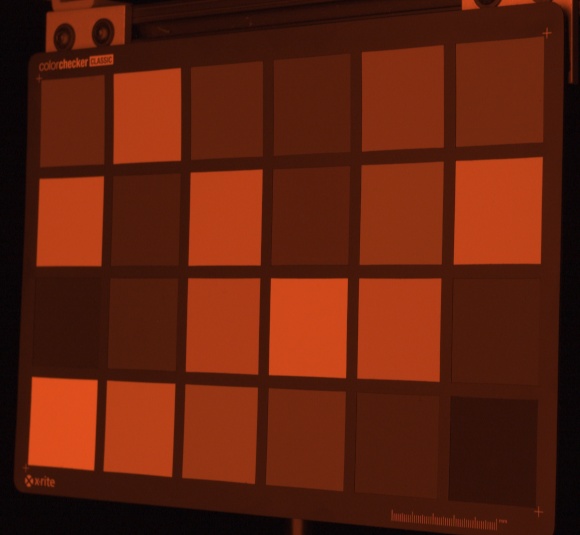} & 
\includegraphics[width=1.in]{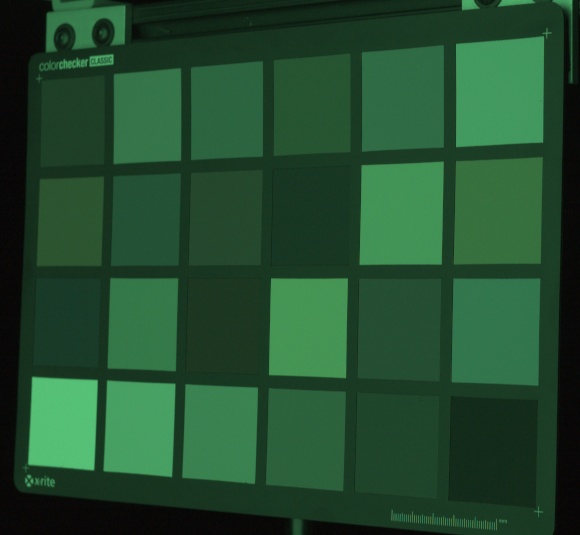} & 
\includegraphics[width=1.in]{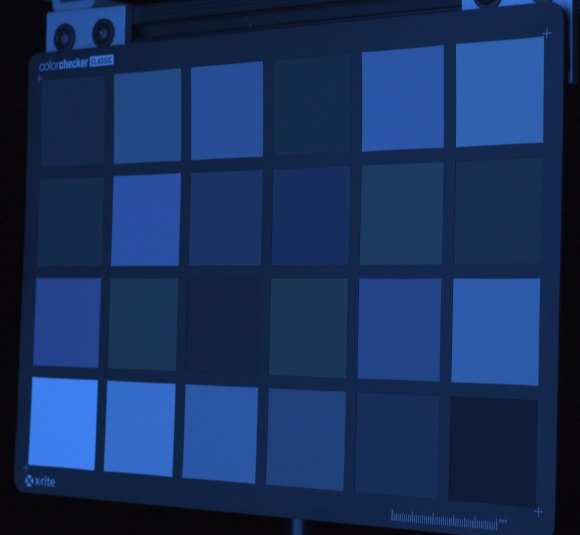} \\
{\rotatebox{90}{\footnotesize{sampled values}}} &
\includegraphics[width=1.in]{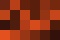} & 
\includegraphics[width=1.in]{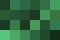} & 
\includegraphics[width=1.in]{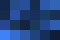} \\

\end{tabular}
\vspace{-5pt}
\caption{Top row: our color rendition calibration setup, photographed with a witness camera. Middle row: calibration photographs of the color chart lit by each spectral channel. Bottom row: sampled from the images of the middle row.}
\label{fig:mrm_calib}
\vspace{-5pt}
\end{figure}

To capture these photographs, we turn on a 1m $\times$ 1m square of the LED wall for each spectral channel, and we place our color chart 1m from the center of the square aimed directly towards the LED wall. We orient our camera such that its optical axis makes a 45$^{\circ}$ angle with the surface normal direction of the color chart, leveraging the color chart's mostly Lambertian reflectance. We show this setup in Fig. \ref{fig:mrm_calib} (top row), and the resulting calibration images in Fig.  \ref{fig:mrm_calib} (middle row), and, finally, the pixel values sampled from these images in Fig. \ref{fig:mrm_calib} (bottom row).

Although we know that sampled values from a color chart illuminated by the VP stage displaying any HDR IBL environment must be a linear combination of the sampled values from the calibration data in Fig. \ref{fig:mrm_calib}, we must determine \textit{how much of each spectral channel} to add together to simulate a color chart's appearance in a given environment. For this, we first need to know the relationship between amount of light represented by the 1m $\times$ 1m square of light as compared with the full sphere of illumination (assuming an ideal virtual production stage with no missing lighting directions). We further need to know how much of each spectral channel will be used when displaying each unique lighting environment.

\begin{figure}[h]
\vspace{-3pt}
\begin{tabular}{c@{ }c}
\includegraphics[width=1.4in]{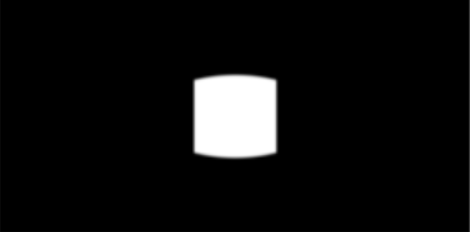} & 
\includegraphics[width=1.4in]{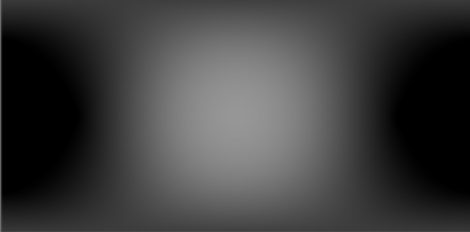} \\
\small{(a) latitude longitude mapping} & \small{(b) diffuse convolution } \\
\end{tabular}
\vspace{-5pt}
\caption{(a) A latitude-longitude mapping of the calibration setup: a 1m $\times$ 1m panel viewed from a 1m distance; (b) The diffuse convolution of (a). From (b) we solve for the scale factor \textbf{$\beta$} that adjusts for the intensity difference between our calibration setup and a full, even sphere of illumination.}
\vspace{-5pt}
\label{fig:square_calib}
\end{figure}

We first define a scale factor \textbf{$\beta$} that accounts for the fact that the 1m $\times$ 1m square of illumination, from a 1m distance, represents a smaller solid angle compared to the full sphere of illumination. We construct a cube map environment with a square representing the panel, using a $\sim$54 pixel width square for a cube map with a face width of 90 pixels ($\sim$54$^{\circ}$). We compute the diffuse convolution for the frontal direction (see Fig. \ref{fig:square_calib}), which yields \textbf{$\beta$} $\approx 0.311$. Thus, we scale our calibration images by $\frac{1}{\beta}$. This scale factor depends only on the setup geometry, and not the individual type of LED panel or camera used. In general, as real-world VP stages will be missing some lighting directions and may not contain emissive flooring, this technique is designed to require only a 1m $\times$ 1m square of LED panels plus a scale factor, rather than omnidirectional, even illumination. Nonetheless, we are still able to predict how a color chart would appear when lit with uniform, omnidirectional illumination for each spectral channel.

Next, to determine how much of each spectral channel will be used when displaying a \textit{particular} lighting environment, we leverage the fact that a color chart reflects light mostly diffusely according to Lambert's law, integrating its full frontal hemisphere of lighting directions. Thus, for the purposes of color rendition, we only care about the diffuse integral of the illumination in the VP stage, rather than the individual contributions of each pixel of the LED panels. We define the diffuse integral of the frontal hemisphere of the HDRI map as the RGB pixel value $\textbf{w}_{\textbf{avg}}$. If the target color chart is photographed while placed at center of projection of the HDRI panorama, then $\textbf{w}_{\textbf{avg}}$ is equal to the pixel value of the white square of this color chart, scaled up to adjust for the fact that the white square of a typical color chart is only $\sim$90\% reflective. Thus, instead of simply summing together images of the color chart as illuminated by a full sphere of illumination for each spectral channel, we can now scale these images based on the expected overall diffuse integral for a given lighting environment, expressed as $\textbf{w}_{\textbf{avg}}$. This is essentially \textit{tinting} the full, even sphere of illumination based on the white balance of the given environment, which can be measured directly from the appearance of the white square of a color chart placed in the original, real-world scene. However, we are employing the pre-correction matrix $\textbf{M}$ when displaying the out-of-frustum content responsible for color rendition. Thus, rather than tinting the full, even sphere of illumination using $\textbf{w}_{\textbf{avg}}$, we must instead tint it using $\textbf{M}\textbf{w}_{\textbf{avg}}$.

Formally, with the above calibration data, we can simulate the appearance of color chart illuminated by an HDR IBL displayed in a VP stage, which we will finally use to solve for $\textbf{Q}$. For a given color chart square $j$, we define a $3 \times 3$ matrix $\textbf{[SRL]}_\textbf{j}$ that encodes the fully-spectral modulation and integration of the camera spectral sensitivity functions, the LED emission spectra, and the material reflectance spectrum. As with the previously defined $\textbf{[SL]}$ matrix, $\textbf{[SRL]}_\textbf{j}$ has the camera's color channels along rows, and the spectral channels of the LED volume along columns. In other words: each column encodes the RGB pixel values of how a particular color chart square $j$ appears when illuminated by each available spectral channel. Note that there will be a different $\textbf{[SRL]}$ matrix for each chart square, as each square has a unique reflectance spectrum. The pixel values for each $\textbf{[SRL]}$ matrix are captured by our calibration process, represented by the sampled values in Fig. \ref{fig:mrm_calib} (bottom row). Finally, the expression to simulate how a given chart square $j$ will appear when illuminated by the VP stage displaying an HDR IBL with diffuse integral $\textbf{w}_{\textbf{avg}}$ and out-of-frustum matrix $\textbf{M}$ is:

\begin{equation}
    \frac{1}{\beta}\textbf{[SRL]}_\textbf{j}\textbf{M}\textbf{w}_\textbf{avg}.
    \label{eqn:SRL}
\end{equation}

Our ultimate goal is to match the color rendition properties of the original scene, with target color chart values $\textbf{p}$. Thus, including the desired post-correction matrix $\textbf{Q}$, we would like to minimize the squared error between the predicted pixel values and the target pixel values across all $n$ chart squares:

\begin{equation}
    \argmin(
    \sum_{j=1}^{n}||\frac{1}{\beta}\textbf{Q}\textbf{[SRL]}_\textbf{j}\textbf{M}\textbf{w}_\textbf{avg} - \textbf{p}_\textbf{j}||).
    \label{eqn:q_solve}
\end{equation}

Each chart square yields three equations (one each for the red, green, and blue channels of the final image), while $\textbf{Q}$ contains nine unknown variables. We could therefore choose three chart squares to match exactly, or we could use all 24 squares of the color chart to solve for  $\textbf{Q}$ in a least squares sense. Recall, in Eqn. \ref{eqn:q_solve}, $\textbf{[SRL]}$ values are obtained from calibration imagery, while  $\textbf{w}_{\textbf{avg}}$ and $\textbf{p}$ are sampled from the target color chart in the original environment. $\textbf{M}$ was computed using the primary-calibration procedure. Thus, while $\textbf{M}$ is lighting environment independent, $\textbf{Q}$ depends on the appearance of a color chart in a particular environment and thus will change depending on the target HDRI map.

\subsubsection*{Solving for $\textbf{N}$: Inverting the effect of $\textbf{Q}$ for the in-camera background}

\begin{figure*}[t]
\vspace{-3pt}
\begin{tabular}{c@{ }c@{ }c@{ }c@{ }c@{ }c@{ }c}
\includegraphics[width=0.95in]{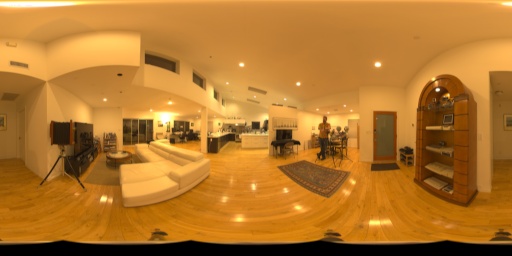} & 
\includegraphics[width=0.95in]{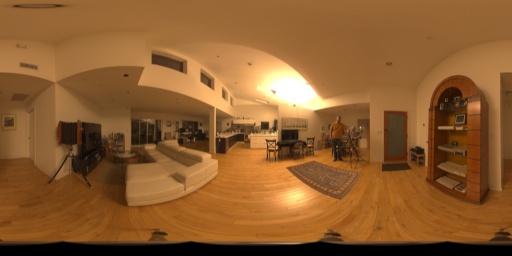} &
\includegraphics[width=0.95in]{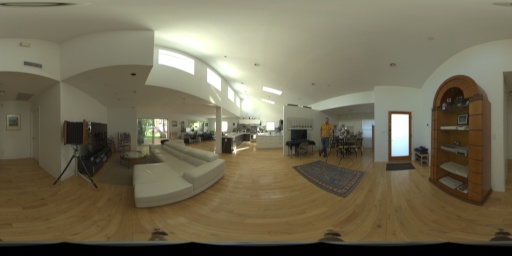} & 
\includegraphics[width=0.95in]{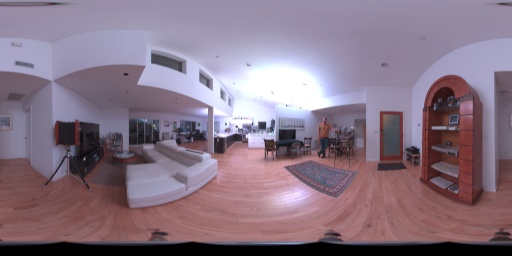} &
\includegraphics[width=0.95in]{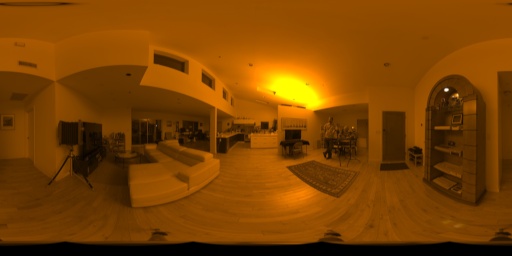} & 
\includegraphics[width=0.95in]{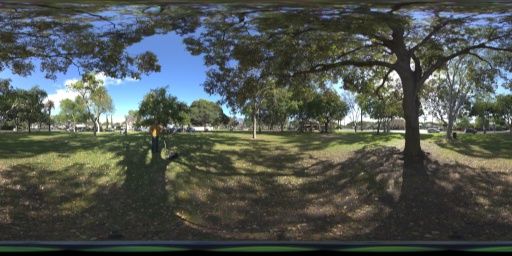} &
\includegraphics[width=0.95in]{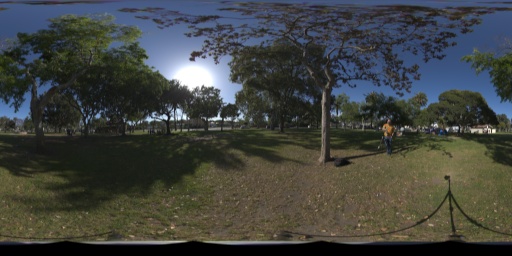} \\
\footnotesize{(1) warm white LED} & \footnotesize{(2) incandescent} &
\footnotesize{(3) indoors daylight} & \footnotesize{(4) RGB LED} &
\footnotesize{(5) sodium vapor} & \footnotesize{(6) outdoors, shade} &
\footnotesize{(7) outdoors, direct sun} \\
\end{tabular}
\vspace{-5pt}
\caption{LDR images of the spectrally-diverse HDR lighting environments that we reproduced in our experiments.}
\vspace{-5pt}
\label{fig:panos}
\end{figure*}

Unfortunately, if we apply the $3 \times 3$ post-correction matrix $\textbf{Q}$ to the whole image, the in-camera background pixels will also be transformed and may no longer appear correct. While foreground detection or rotoscoping could allow the correction to be applied only to the foreground, this would require additional complexity.  Our key insight here is that if we know in advance the post-correction matrix $\textbf{Q}$ that we will ultimately apply to the final image or video sequence, we can apply the inverse of this correction to the in-frustum part of the LED panels with $\textbf{Q}^{-1}$. As the background pixels do not contribute significantly to the lighting on the actors, both the color rendition on the actors and the appearance of the in-camera pixels can be optimized at the same time, with no foreground/background separation required. Light reflecting at grazing angles from the background to the camera generally has less opportunity to interact with the pigments and chromophores of a material — with Fresnel-enhanced specular reflections from dielectric materials such as skin being an extreme case — and thus will tend to maintain both the color and spectrum of the original illumination. As such, we expect color rendition errors from the background lighting to be relatively unnoticeable.

Given $\textbf{Q}$ and $\textbf{SL}$, we rewrite Eqn. \ref{eqn:m_solve0} to solve for a new in-camera-frustum pre-correction matrix $\textbf{N}$:
\vspace{-1pt}
\begin{equation}
    \textbf{Q}\textbf{[SL]}\textbf{NI} = \textbf{I}.
\end{equation}

\vspace{-5pt}
\begin{equation}
     \textbf{N} =  \textbf{[SL]}^{-1}\textbf{Q}^{-1}.
\end{equation}

Furthermore, we can also substitute $\textbf{M}$ for $\textbf{[SL]}^{-1}$, clarifying the relationship between the in- and out-of-camera-frustum matrices:
\vspace{-2pt}
\begin{equation}
    \textbf{N} = \textbf{M}\textbf{Q}^{-1}.
\end{equation}

In summary, we pre-correct the in-frustum content with $\textbf{N}$, expecting to post-correct the footage with $\textbf{Q}$. As $\textbf{Q}$ is lighting environment dependent, so is $\textbf{N}$.  In practice, since $\textbf{Q}$ will typically desaturate pixel colors, $\textbf{Q}^{-1}$ will typically increase the saturation of pixel colors. As such, we ensure that the LED panels are set to display in their widest possible color gamut, with primaries that turn on each LED spectrum independently.

\subsection{Black Level Subtraction}
A current limitation of LED panels used for in-camera backgrounds is that the panels themselves reflect back some incident illumination, i.e., they have a non-zero \textit{albedo}. Thus LED panels lighting the actors also have the unwanted side-effect of illuminating the in-camera background panels. In our experiments, we observed panel albedos varying from $\sim$4-10\%. To compensate for this, we adjust the in-camera-frustum pixel values with an RGB offset using a "black level" measurement. For each lighting environment, we turn on all the out-of-frustum content, first applying $\textbf{M}$. Next, we turn off the in-camera background, and record with our camera a per-lighting-environment average RGB color to subtract from the rendered content prior to display. In practice, we sample a region of interest from such an image, which yields the RGB pixel value $b_{camera}$. This pixel value cannot be used directly for the black level subtraction for the content, as we need to factor in the scale factor between the content and its camera-observed value. We can sum together the calibration images of Fig. \ref{fig:mim_calib} to obtain a pixel value $w_{camera}$ of the camera observing a pixel value of $[1, 1, 1]$. We then compute the black level to subtract from the final rendered content as $\frac{b_{camera}}{w_{camera}}$. The downside of this approach is that it requires per-lighting-environment empirical measurement. Future work could aim to analytically estimate the black level subtraction via light transport, given $w_{camera}$, measured panel albedo, and the HDRI map, or to measure the radiosity form factors within the stage directly using a wide-angle camera and a discretized lighting basis.

\section{Results}

In this section, we present experimental and theoretical results, evaluating our technique when reproducing the color rendition properties of several real-world lighting environments.

\subsubsection*{HDR Lighting Measurement and Real-World Photography}
We captured seven spectrally-diverse lighting environments using a nodal tripod mount, Canon 5D Mark III, and a Canon 8-15mm fisheye lens with a 180$^\circ$ field-of-view, using multiple exposure HDR photography. Five of these lighting environments were captured indoors in a living room, each with its own distinct type of illumination: (1) broad-spectrum warm white LED lighting, (2) incandescent (tungsten halogen) lighting, (3) daylight shining through windows, (4) RGB LED "white" light, and (5) monochromatic yellow-orange sodium vapor lighting, with a distinct emission spectrum spike at 589 nm. The remaining two lighting environments were captured outdoors at a park, (6) in the shade and (7) in direct sunlight. We show LDR renderings of the lighting environments in Fig. \ref{fig:panos}. We were careful that no light sources clipped at each shortest exposure, with the exception of the sun, whose missing energy we reconstructed using the appearance of the white square of a color chart placed in the scene leveraging prior techniques \cite{Debevec:2003}. We calibrated an aperture-dependent radial lens fall-off, appropriately scaling the merged HDR images before stitching together four different views to assemble each lighting environment into an 8k panorama with PTGui \shortcite{PTGui}.

In each environment, to capture the color rendition properties of the real-world illumination, we additionally photographed two subjects wearing brightly colored patterned clothing, along with a color chart, a diffuse gray sphere, and a mirrored sphere as reference. For these images, we ensured that the panorama's center of projection was located near the placement of the color chart in the scene. For these reference photographs, we used the same camera outfitted with a 40mm lens and an HDR exposure series matching that of the panorama capture.

\subsubsection*{Lighting Reproduction in an RGB VP Stage}
To test our technique, we reproduced each lighting environment inside a cylinder-shaped RGB LED based virtual production stage including floor, ceiling, and wall LED panels. Although the stage did not produce a full 360$^\circ$ of illumination, we ensured a setup that maximized coverage of frontal lighting directions while still allowing us to record an in-camera background. Inside the VP stage, we captured all images using a ZCam E2 compact cinema camera outfitted with a Laowa 17mm MFT lens. Our spherical content was displayed on the LED panels using the Unreal Engine, with camera tracking to adjust the camera frustum in real-time. Within Unreal, we added a color matrix operation to the in-frustum and out-of-frustum shaders separately to implement our approach.

To evaluate the effect of $\textbf{Q}$ and $\textbf{N}$ on the in-camera background displayed colors, we composited the color charts photographed in the real environments into the HDRI maps, careful to match the exposures, such that they would appear behind our actors when displaying the HDRI maps. As our main focus was color rendition, we did not try to match the perspective of the virtual camera with our real world camera, so the in-camera backgrounds in our experimental results are not precisely aligned with those of the real-world imagery. The backgrounds displayed in the VP stage are from the HDRI map, shot from the perspective of the subjects rather than from that of the camera used for our reference portraits.

In general, real-world light sources will be orders of magnitude brighter than the rest of a scene. However, in practice, VP stages have a limited dynamic range. To achieve a sufficiently bright in-camera background appearance for the composited color charts, we required a white square pixel value greater than 0.1. With this constraint, the maximum pixel values of the HDRI maps ranged from 40 to 3000 (allowing sun energy to spread to more than one pixel), well beyond the maximum panel-displayable value of 1. To prevent light source clipping in the VP stage, prior to displaying the content we used an energy-preserving light source dilation algorithm \cite{Debevec:2022} for each HDRI map.

\subsubsection*{Experimental Results}
In this section, we use one consistent color matrix to take all images from a camera raw color space to sRGB for display as a post-process for the purpose of visualization. This matrix was computed from a color chart photographed in daylight using the Canon 5D Mark III. When we report quantitative error, however, these metrics are computed in the camera raw color space.

We also present color charts with inset circles, where the background squares represent pixel values sampled from a color chart photographed in a real-world scene, while foreground circles represent values sampled from a chart in the VP stage. We sample chart values both from a color chart \textit{illuminated} by out-of-frustum content and one \textit{displayed} on the LED panels in-frustum and observed directly by the camera. For these visualizations, we scale the sampled pixel values such that the green channel of the white squares match to facilitate visual comparisons. In practice, our VP stage is missing some lighting directions, so the out-of-frustum lit chart is typically dimmer than the in-frustum displayed chart.

In Fig. \ref{fig:fullprocess}, we demonstrate our full process including intermediate results for the outdoors, shaded environment. As compared with the real photograph captured outdoors [column (a)], reproducing the illumination in the VP stage using the baseline primary-based calibration approach leads to the expected overly saturated skin tones and reddish hue shift for yellow and orange materials [column (b)]. Applying the post-correction matrix $\textbf{Q}$ [column (c)] desaturates these colors, improving color rendition, as observed for the comparison charts, skin tones, and clothing. However, $\textbf{Q}$ also adds a blueish tint to the in-camera \textit{displayed} chart, visible in the comparison chart of column (c). Applying $\textbf{N}=\textbf{M}\textbf{Q}^{-1}$ to the in-frustum content removes this blueish tint [column (d)], but color rendition is still poor for the in-camera background due to the non-zero panel albedo and bounced light. The black level subtraction for the in-frustum content [column (e)] thus provides the most dramatic improvement for the in-camera background.  While small mismatches remain, we achieved our target goal of desaturating skin tones and improving orange/yellow material color rendition, without sacrificing the in-camera background colors. Unfortunately, the VP stage lacked light sources for the directions corresponding to those providing the rim lighting on the subject's left side of the face, somewhat limiting the overall quality of the reproduced illumination.

\input{overall_process_figure0.tex}
\input{all_results_figure1.tex}
\input{all_results_figure2.tex}

For the remaining six environments, we show the real-world photographs, then the baseline VP lighting reproduction using $\textbf{M}$ only for both in- and out-of-frustum content, and then our full multi-matrix pipeline with black level subtraction in Fig. \ref{fig:all_env_results1} and Fig. \ref{fig:all_env_results2}. When using the baseline calibration approach where the goal is just to match the in-camera background colors, the \textit{lit} color chart in the VP stage is overly saturated, with hue shifts in yellow and orange materials. One exception is the RGB LED based white lighting environment, which the VP stage is able to reproduce quite well owing to the spectral similarity between its light sources and those of this particular real-world scene. For this scene, the post-correction $\textbf{Q}$ matrix is close to the identity and does not have a very significant effect. However, there are color rendition challenges for the remaining scenes that include broader spectrum illumination sources. For the remaining scenes, the $\textbf{Q}$ matrix is able to desaturate colors as required, leading to improved color rendition for the \textit{lit} chart. While $\textbf{Q}$ desaturates the overall image content, the in-frustum matrix $\textbf{N}$ ensures that the in-camera background content is still as close as possible to the target color appearance. However, as in Fig. \ref{fig:fullprocess}, the black level subtraction for panel albedo is the more significant effect for in-camera background color rendition.

In the most extreme case, in Fig. \ref{fig:all_env_results2} (upper row), we show that the RGB LEDs are not \textit{at all} able to reproduce the narrow-band illumination of the sodium vapor light source. While under the real sodium vapor lighting, materials appear nearly monochromatic (tinted yellow), under the RGB-reproduced illumination they still have discernible colors. The post-correction matrix $\textbf{Q}$ completely desaturates these colors, satisfyingly reproducing the monochromatic look of the original scene. However, there is a noticeable difference between the subjects' clothing in the real-world scene compared with in the VP stage, even after applying $\textbf{Q}$: the yellow regions of the clothing remain too dark. In this case, a $3 \times 3$ post-correction matrix can only help so much. Furthermore, for the sodium vapor environment, we were unable to solve for $\textbf{N}$. $\textbf{Q}$ was ill-conditioned due to the monochromatic nature of the illumination and thus $\textbf{Q}^{-1}$ is poorly defined. As such, for Fig. \ref{fig:all_env_results2} (upper row), our full technique includes only $\textbf{M}$, $\textbf{Q}$, and the black level subtraction.

Unexpectedly, we observed some remaining differences in the white balance of the \textit{lit} color charts and the \textit{displayed} color charts, despite our best attempts to calibrate the full system. As an example, see the results of Fig. \ref{fig:all_env_results1} (bottom row). While we did our best to ensure panel linearity, we observed the red LED channel of our VP stage exhibited its own single-channel non-linearity. We suspect that a single-channel linearity calibration process involving a 3D lookup table (LUT) could mitigate this issue. 

\subsubsection*{Quantitative Results} 
\begin{figure}[h]
\vspace{-4pt}
\begin{tabular}{c}
\includegraphics[width=3in]{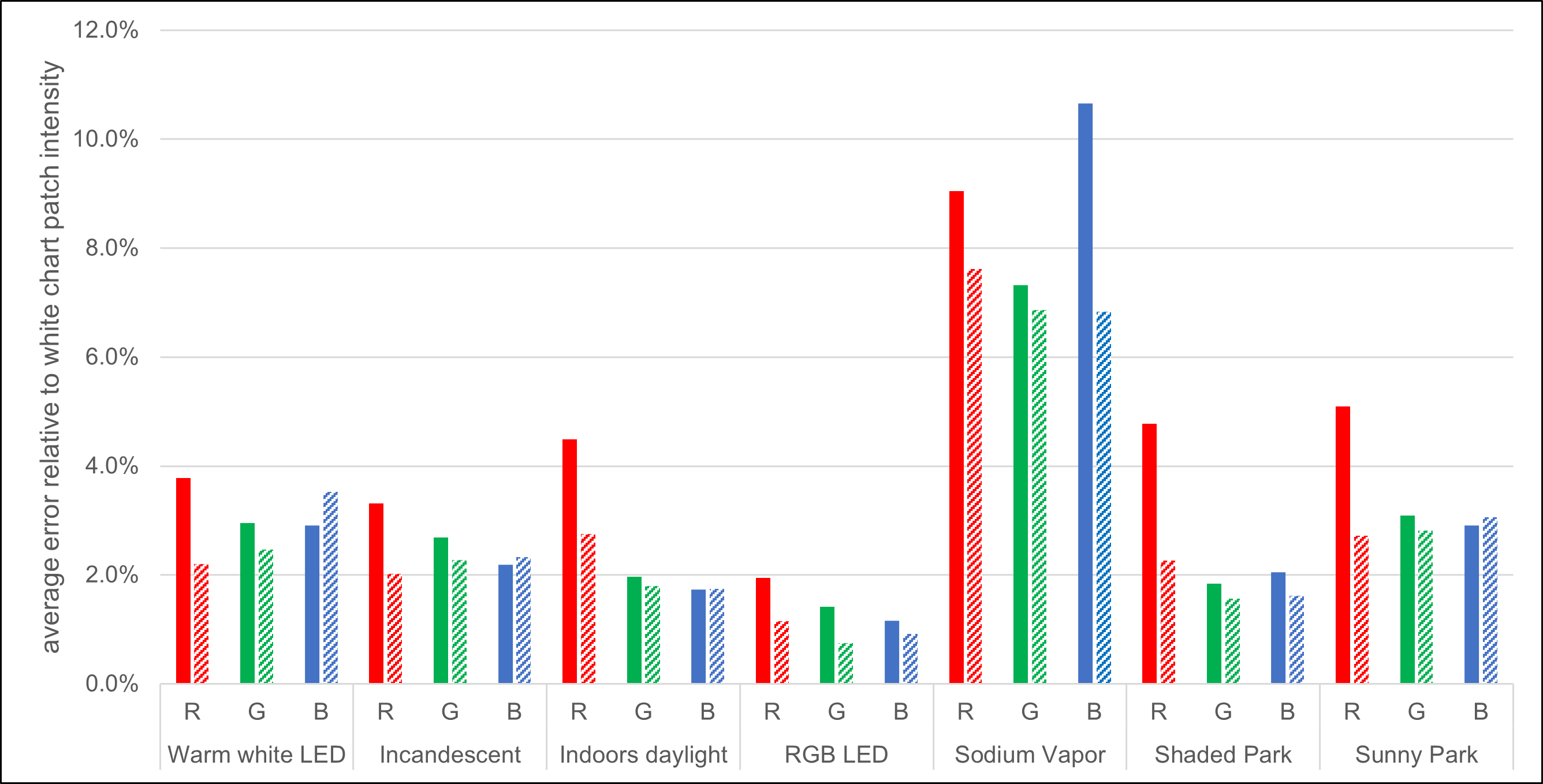} \\
(a) out-of-frustum \textit{lit} color chart error \\
\includegraphics[width=3in]{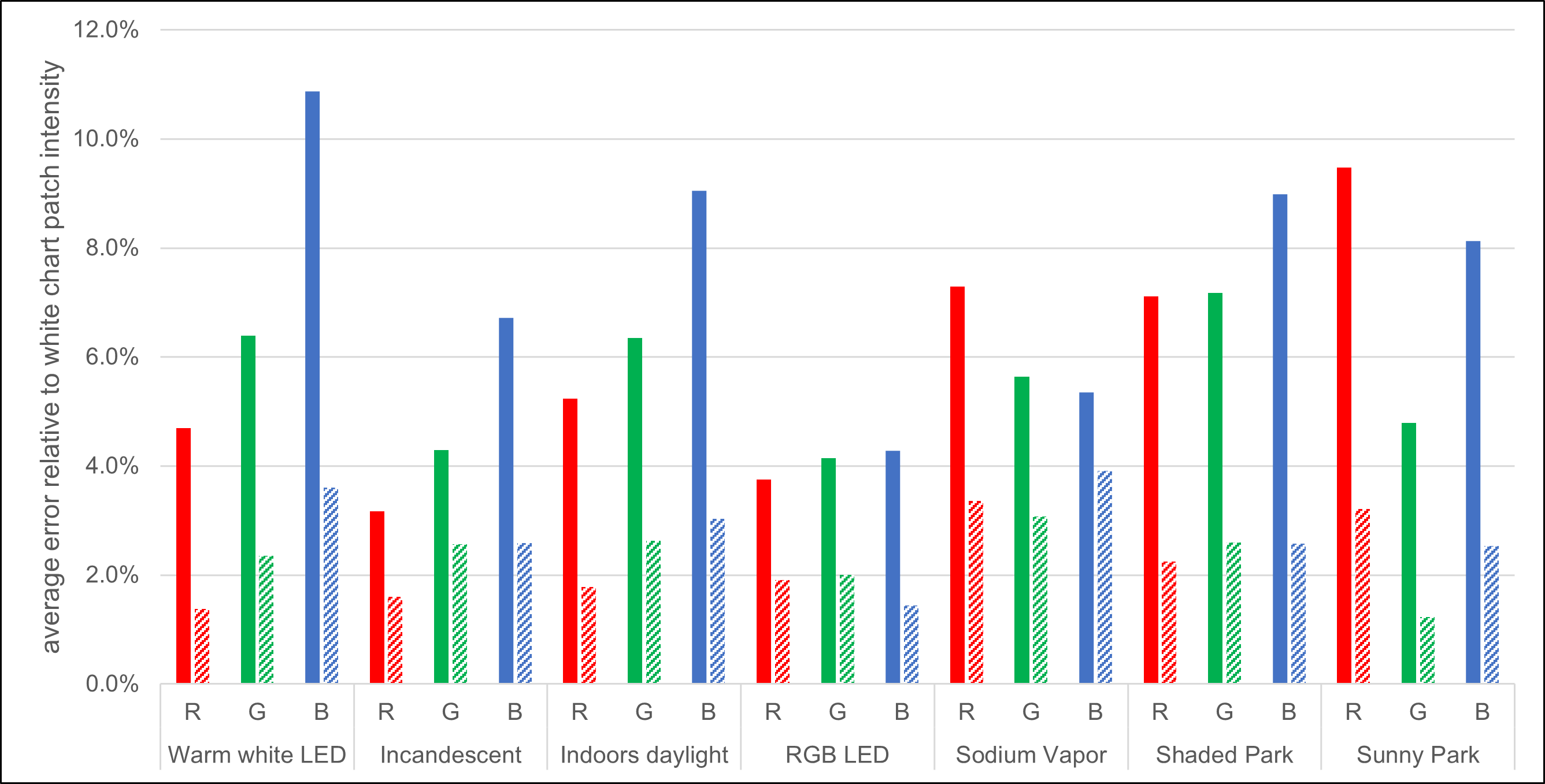} \\
(b) in-camera-frustum \textit{displayed} color chart error \\
\end{tabular}
\vspace{-6pt}
\caption{Average error relative to the white chart patch intensity value, for RGB channels individually and for each lighting environment, (a) for the out-of-frustum \textit{lit} color chart, compared with the target color chart in the real world illumination, and (b) for the in-frustum \textit{displayed} chart, also compared with the target. The solid bars represent error for the baseline approach ($\textbf{M}$ only) while the dashed bars represent error for our full approach leveraging $\textbf{Q}$ and, for the displayed chart, black level subtraction.}
\vspace{-7pt}
\label{fig:quant0}
\end{figure}

In Fig. \ref{fig:quant0} we show quantitative errors corresponding to the comparison color charts of Figs. \ref{fig:all_env_results1} and \ref{fig:all_env_results2}. Solid bars represent the per-color-channel error associated with the baseline approach, while dashed bars represent error for our full approach. For both the out-of-frustum \textit{lit} [Fig. \ref{fig:quant0}(a)] and in-camera-frustum \textit{displayed} [Fig. \ref{fig:quant0}(b)] color charts, the average error relative to the white square intensity is under 4\% for all seven scenes. The RGB LED lighting environment has the smallest average error, while the sodium vapor has the largest error for the \textit{lit} color chart, consistent with visual observations from Figs. \ref{fig:all_env_results1} and \ref{fig:all_env_results2}.

\subsubsection*{Theoretical Results}
\begin{figure}[h]
\vspace{-3pt}
\begin{tabular}{@{ }c@{ }c@{ }c@{ }c@{ }c@{ }}
{\rotatebox{90}{\footnotesize{(1)}}} &
\includegraphics[width=0.75in]{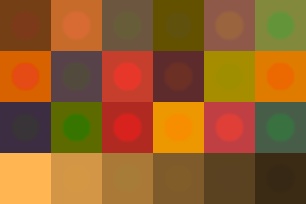} & 
\includegraphics[width=0.75in]{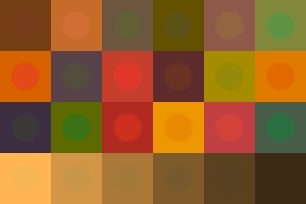} & 
\includegraphics[width=0.75in]{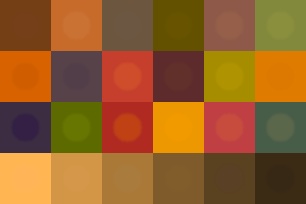} & 
\includegraphics[width=0.75in]{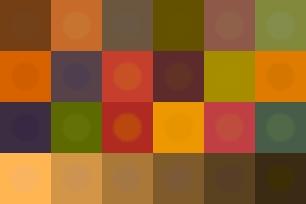} \\
{\rotatebox{90}{\footnotesize{(2)}}} &
\includegraphics[width=0.75in]{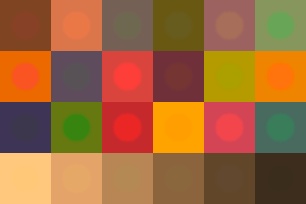} & 
\includegraphics[width=0.75in]{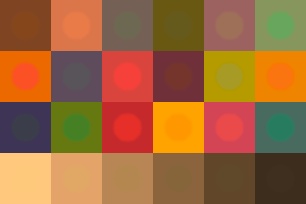} & 
\includegraphics[width=0.75in]{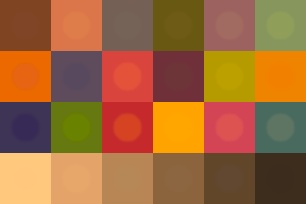} & 
\includegraphics[width=0.75in]{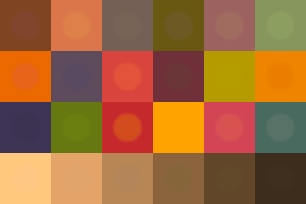} \\
{\rotatebox{90}{\footnotesize{(3)}}} &
\includegraphics[width=0.75in]{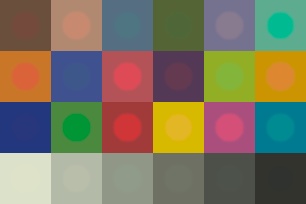} & 
\includegraphics[width=0.75in]{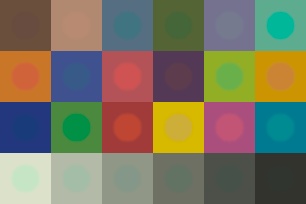} & 
\includegraphics[width=0.75in]{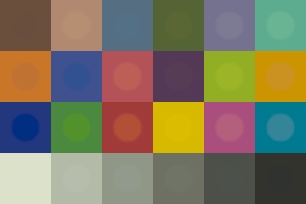} & 
\includegraphics[width=0.75in]{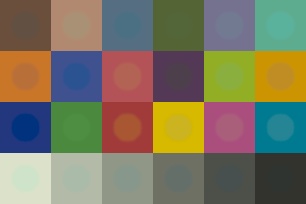} \\
{\rotatebox{90}{\footnotesize{(4)}}} &
\includegraphics[width=0.75in]{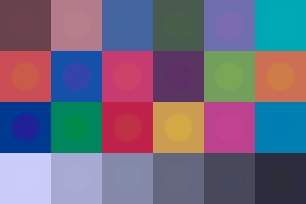} & 
\includegraphics[width=0.75in]{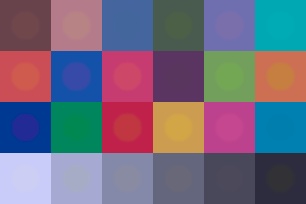} & 
\includegraphics[width=0.75in]{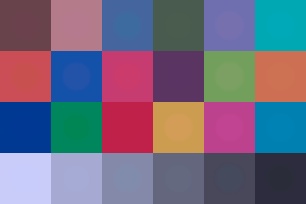} & 
\includegraphics[width=0.75in]{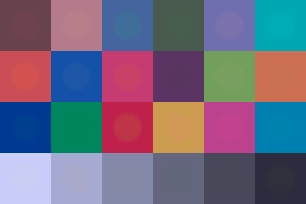} \\
{\rotatebox{90}{\footnotesize{(5)}}} &
\includegraphics[width=0.75in]{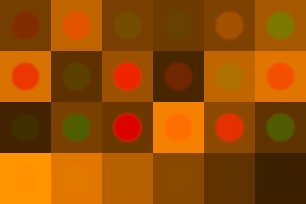} & 
\includegraphics[width=0.75in]{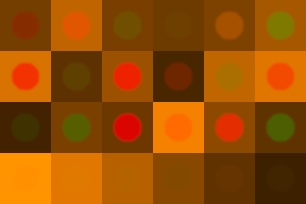} & 
\includegraphics[width=0.75in]{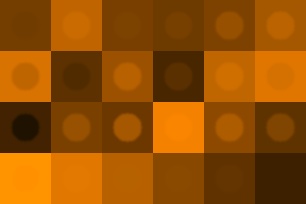} & 
\includegraphics[width=0.75in]{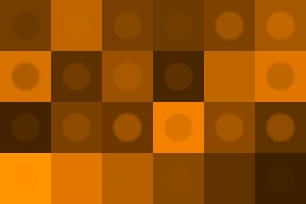} \\
{\rotatebox{90}{\footnotesize{(6)}}} &
\includegraphics[width=0.75in]{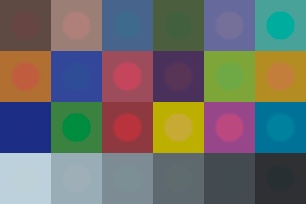} & 
\includegraphics[width=0.75in]{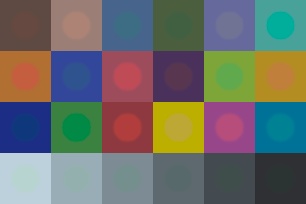} & 
\includegraphics[width=0.75in]{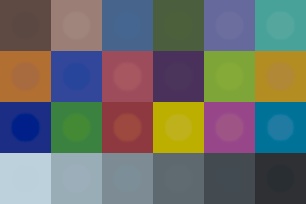} & 
\includegraphics[width=0.75in]{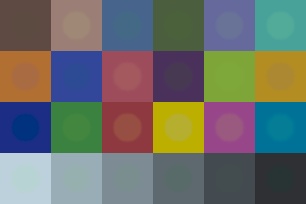} \\
{\rotatebox{90}{\footnotesize{(7)}}} &
\includegraphics[width=0.75in]{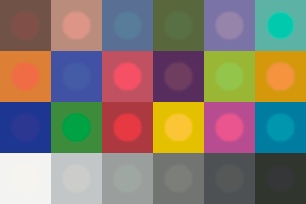} & 
\includegraphics[width=0.75in]{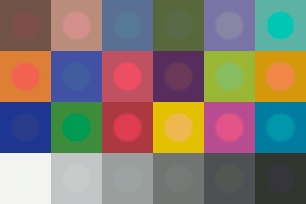} & 
\includegraphics[width=0.75in]{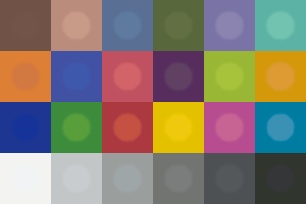} & 
\includegraphics[width=0.75in]{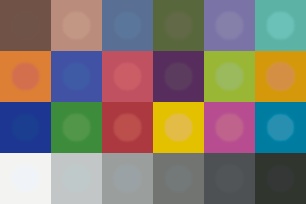} \\
&
\footnotesize{(a) $\textbf{M}$ only (theory)} &
\footnotesize{(b) $\textbf{M}$ only (exp.)} &
\footnotesize{(c) $\textbf{M}$ \& $\textbf{Q}$ (theory)} &
\footnotesize{(d) $\textbf{M}$ \& $\textbf{Q}$ (exp.)} \\
\end{tabular}
\vspace{-5pt}
\caption{Comparison charts demonstrating the difference between our technique in theory and in experimental practice. Background squares are sampled from the chart in the real environment, while foreground circles are sampled from corresponding charts for each column label. (a) shows using $\textbf{M}$ only for the \textit{lit} chart, a theoretical result computed using Eqn. \ref{eqn:SRL}. (b) shows using $\textbf{M}$ only, with pixel values sampled from images captured in the VP stage for the \textit{lit} chart during our experiments. (c) shows adding $\textbf{Q}$ to the theoretical result of (a). (d) shows adding $\textbf{Q}$ to the experimental result of (b). Our experimental and theoretical results match closely. The lighting environments for each row are indicated at the left, using the labels of Fig. \ref{fig:panos}.}
\vspace{-5pt}
\label{fig:theory_vs_practice}
\end{figure}

Although we have demonstrated the practical performance of our technique, for each lighting environment, we also show theoretical results for the \textit{lit} color charts in Fig. \ref{fig:theory_vs_practice}. Rather than record the color chart's appearance, as in the previous section, here we also simulate the appearance of the color chart using Eqn. \ref{eqn:SRL} and our calibration data, first using $\textbf{M}$ only and then using $\textbf{M}$ and $\textbf{Q}$. Our experimental results using $\textbf{Q}$ [Fig. \ref{fig:theory_vs_practice}(d)] are very similar to the theoretical results using $\textbf{Q}$ [Fig. \ref{fig:theory_vs_practice} (c)]. The remaining color error in [Fig. \ref{fig:theory_vs_practice}(d)], when our results are compared with the target color chart background squares, is therefore largely not the product of calibration or real-world system errors, but, rather, a fundamental limitation of using only $3 \times 3$ linear transformations to improve color rendition from RGB LED based illumination.

\section{Future Work}
\subsubsection*{Evaluation for Different Skin Tones} While our technique improves color rendition for the squares of a color chart and several colorful fabrics, and it desaturates the overly-pink appearance of lighter skin tones in VP stages, future work should also evaluate its performance when photographing a greater diversity of subjects with different skin tones, as in Kadner \shortcite{Kadner:2021a} and LeGendre et al. \shortcite{LeGendre:2016}.

\subsubsection*{Joint Optimization of $\textbf{M}$ and $\textbf{Q}$} We have referred to our multi-matrix approach as \textit{near optimal} rather than \textit{optimal}, as we have first fixed the out-of-frustum matrix $\textbf{M}$ and then subsequently solved for the post-correction matrix $\textbf{Q}$. Upon closer inspection, one could imagine minimizing the same objective function as in Eqn. \ref{eqn:q_solve}, but jointly optimizing for $\textbf{M}$ and $\textbf{Q}$ simultaneously. Initially, we did try this joint optimization approach, but we found in practice that the resultant illumination in the VP stage appeared very non-neutral in color, and the post-correction matrix $\textbf{Q}$ often had to make dramatic color adjustments. Beyond the obvious issue of such lighting being non-ideal to actors immersed in the content of the VP stage, this optimization approach often led to practical out-of-gamut issues in the VP stage. Future work could explore this joint optimization approach, while constraining the white point of the content for better on set appearance and in-gamut content.

In our current approach, we have also not addressed the ideal color space of the content to be displayed. In our experiments, our HDRI maps were encoded in the camera raw of the Canon 5D Mark III camera, a relatively desaturated color space. This color space was somewhat sensible, as the majority of the displayed content was well within the achievable color gamut of the LED panels. However, a joint optimization approach could allow us to further reason about the ideal color space for the content to be displayed.

\subsubsection*{Moving Beyond Linear Color Transforms with Machine Learning} The theoretical results of Fig. \ref{fig:theory_vs_practice} demonstrate the limits of our proposed technique when leveraging $3\times3$ color matrix transforms. The lighting reproduced using RGB LEDs is fundamentally lacking energy in parts of the visible spectrum, so only so much color information can be recovered. Future work could leverage exemplar-based machine learning techniques to effectively hallucinate the appearance of materials under these missing parts of the EM spectrum based on multispectral image datasets or training data captured under both RGB and broad-spectrum illumination, taking inspiration from recent successes in exemplar-based gray-scale image colorization (e.g. \cite{Zhang:2016}).

\subsubsection*{Mixing Practical Lights and Virtual Production} On typical virtual production sets, cinematographers may also wish to add practical light sources into the scene to enhance the look and feel of a particular shot. In our work, we have not yet addressed how to drive the LED panels and external practical lights \textit{together} for optimal color rendition, which we view as an additional opportunity for future work. We imagine that our current framework, however, may still prove useful for this sub-problem.

\subsubsection*{Extending our Approach to CG Scenes} Finally, although we have demonstrated the results of our technique using real-world photographed HDR IBL environments, it should apply as well when using rendered or computer-generated (CG) HDR IBL environments. However, one requirement of our technique is that we know the color rendition of the target lighting environment, provided via a color chart photographed in the original scene. One could either render such a color chart for a CG scene using a spectral rendering technique, or, alternatively if such a renderer is unavailable one could simply capture a photograph of a color chart in a similar real-world scene. For example, for a CG daylight environment, one could photograph a color chart in the real world in daylight as a record of the target color rendition. Future work could evaluate this proposed technique for bridging the gap between rendered RGB HDR IBL environments and real-world measured color rendition.

\section{Conclusion}

We have presented a novel technique to improve the color rendition for RGB-LED based lighting reproduction, as applied towards today's LED panel virtual production stages. In our approach, we treat the primary goal as improving color rendition for materials illuminated by out-of-camera-frustum content in the LED volume, while treating in-camera color rendition as a secondary goal. Our technique requires only four calibration images: a record of how each LED appears to the motion picture camera and how each spectral channel of the RGB LED volume lights a color chart. We derive from these four images and a target color chart appearance three separate color transforms represented as $3\times3$ matrices: one that corrects in-frustum content, one that corrects out-of-frustum content, and one that is applied as a post-process to the acquired footage. We demonstrated that our method out-performs the previous state-of-the-art technique for color calibration, which only calibrates for the in-camera background color appearance. Our technique is straightforward, requiring only a few additional calibration images and basic shaders applied in a virtual production system.

\begin{acks}
We thank Connie Siu and Stephan Trojansky for facilitating our VP stage capture session.
\end{acks}

\bibliographystyle{ACM-Reference-Format}
\bibliography{digipro}

\end{document}

%% file: overall_process_figure0.tex
\begin{figure*}[p]
\vspace{-3pt}
\begin{tabular}{r@{ }r@{ }c@{ }c@{ }c@{ }c}

{\rotatebox{90}{\footnotesize{photograph}}} &
\includegraphics[width=1.3in]{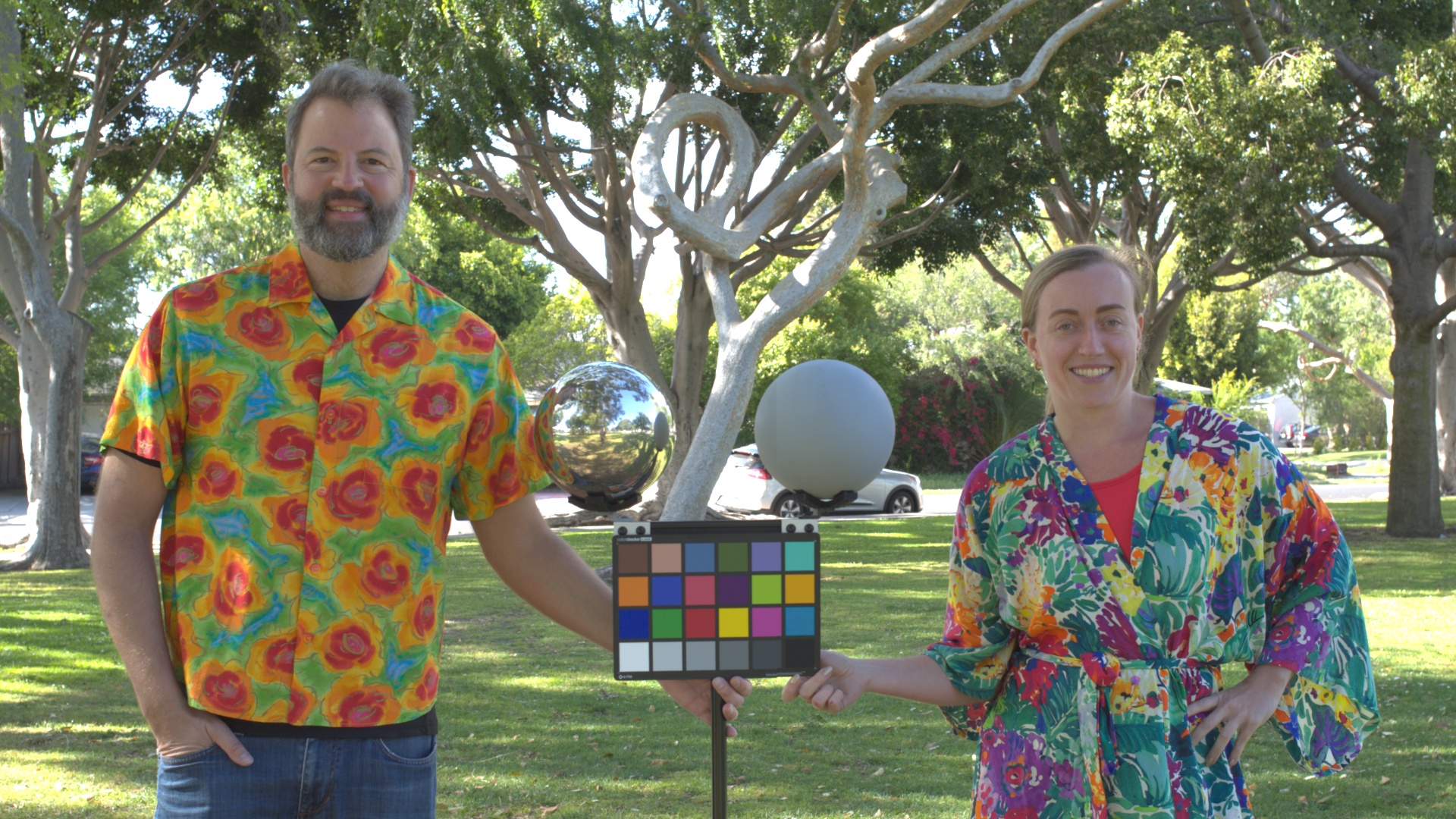}&
\includegraphics[width=1.3in]{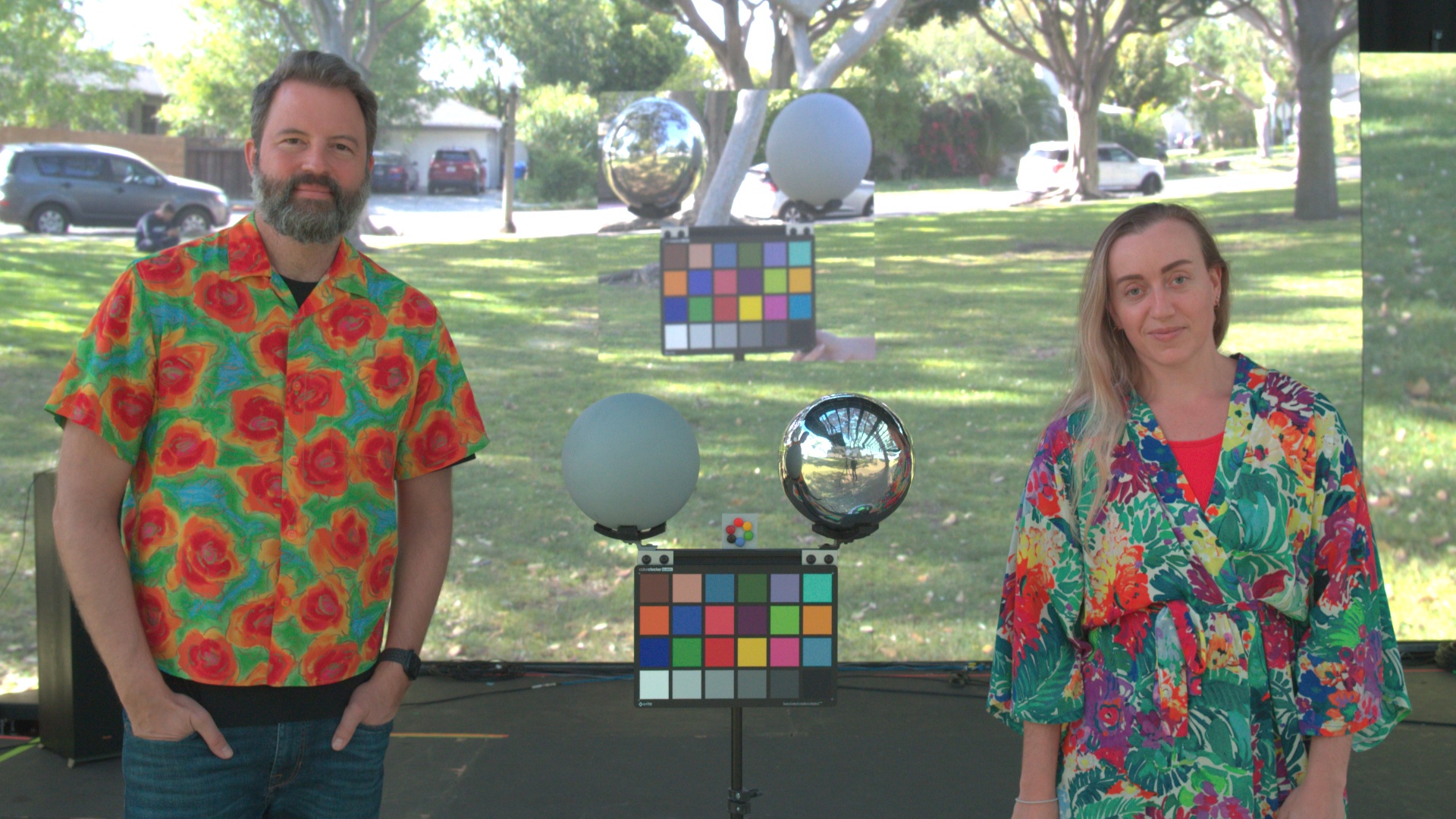} &
\includegraphics[width=1.3in]{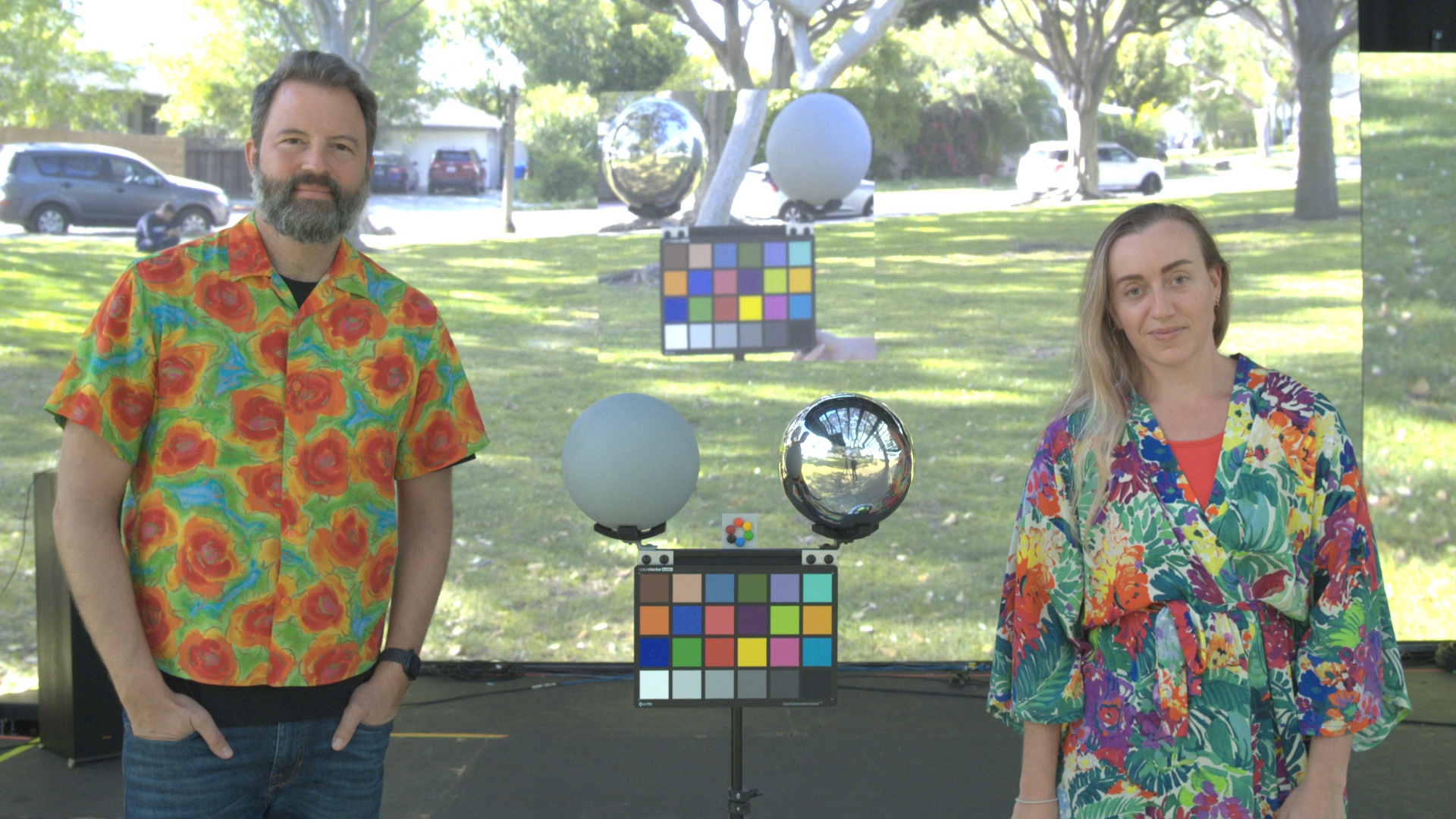} &
\includegraphics[width=1.3in]{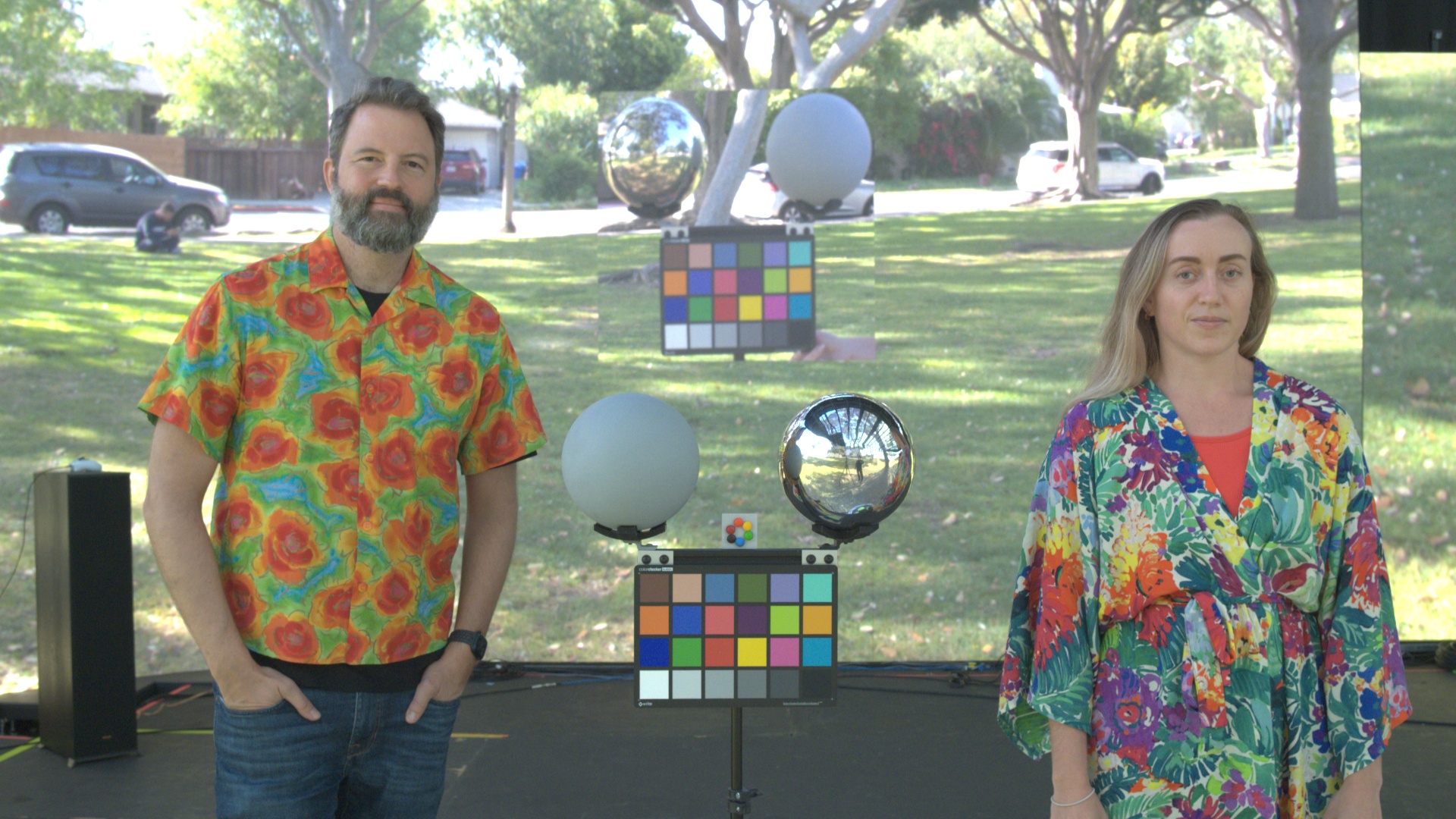} &
\includegraphics[width=1.3in]{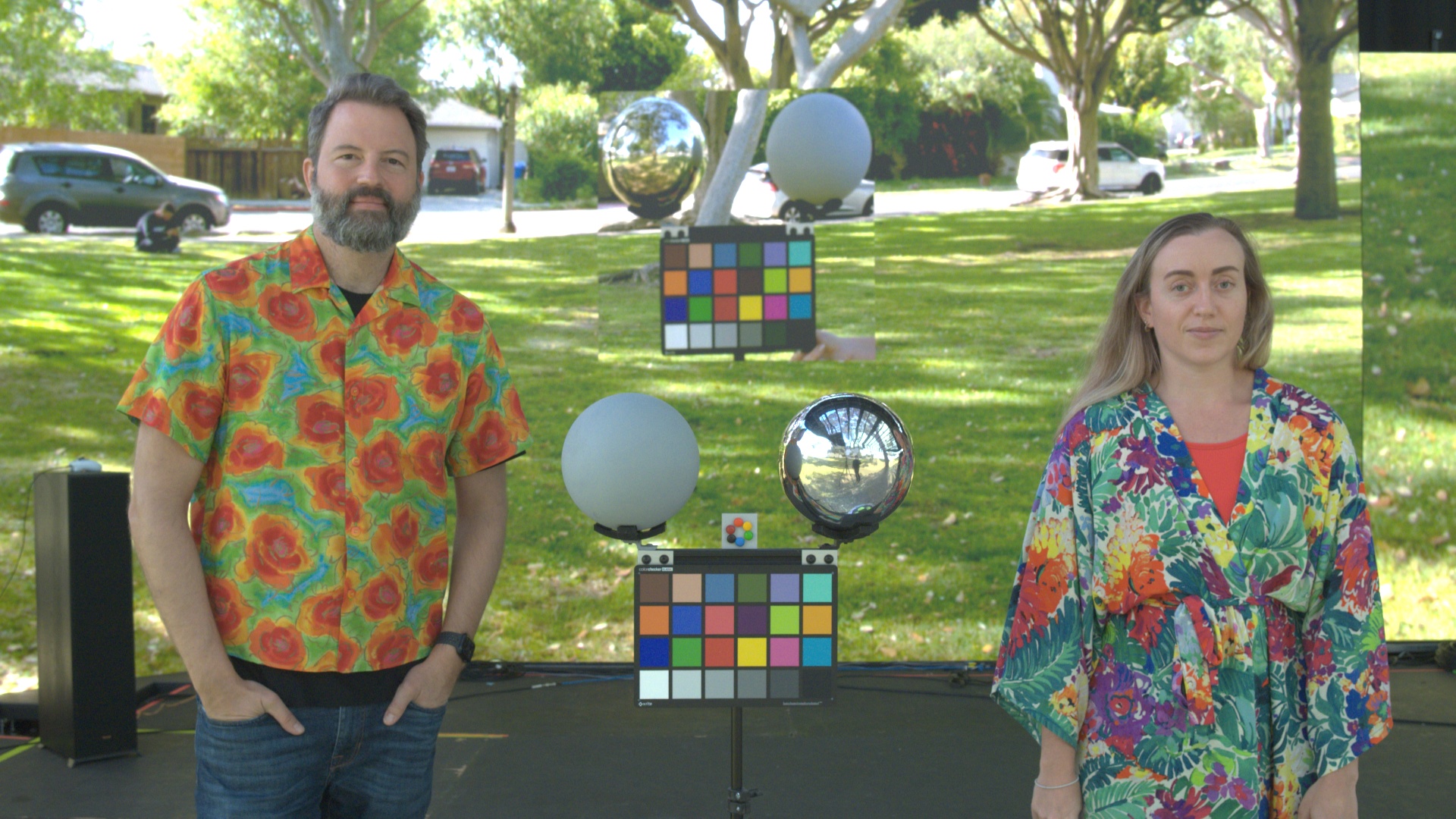}\\

{\rotatebox{90}{\footnotesize{crop of patterned shirt}}} &
\includegraphics[width=1.3in]{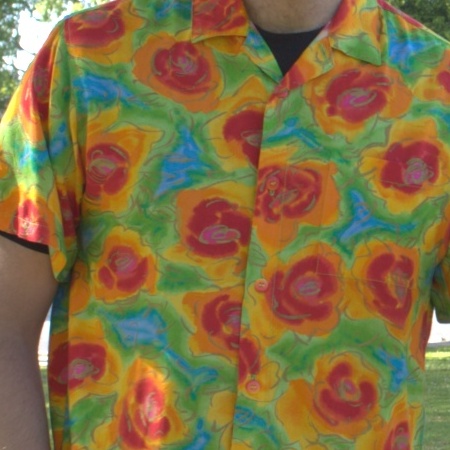}&
\includegraphics[width=1.3in]{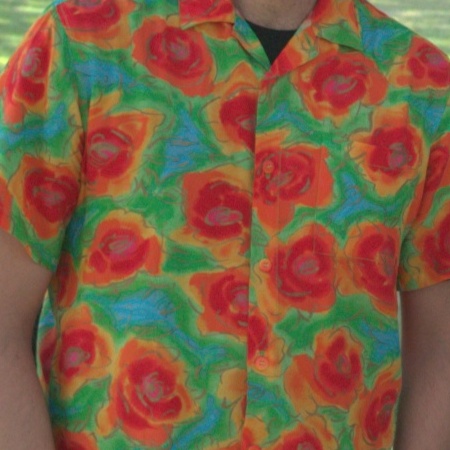} &
\includegraphics[width=1.3in]{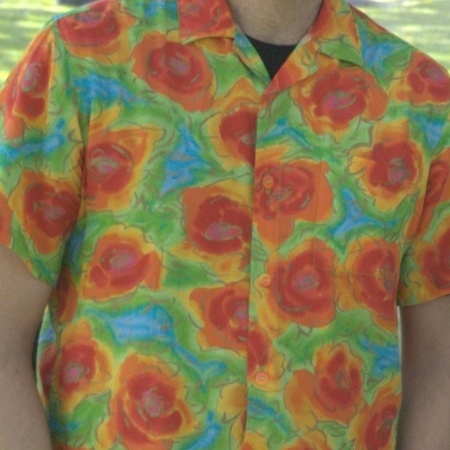} &
\includegraphics[width=1.3in]{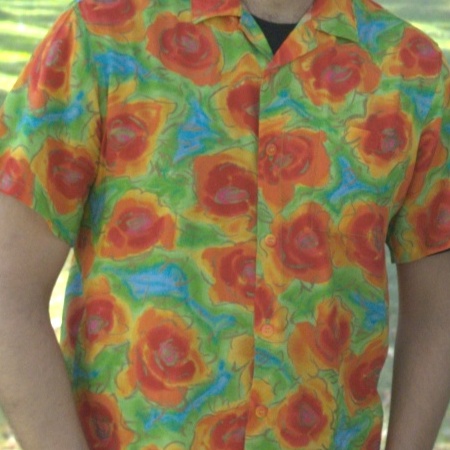} &
\includegraphics[width=1.3in]{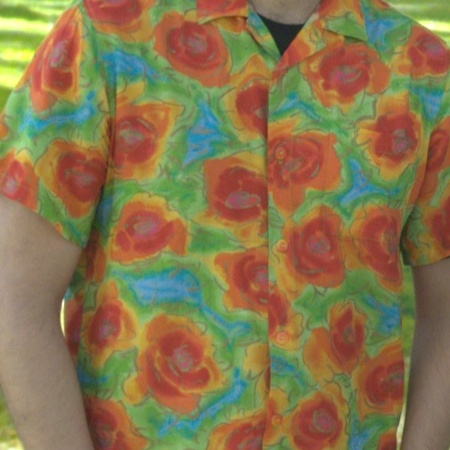}\\

{\rotatebox{90}{\footnotesize{crop of face}}} &
\includegraphics[width=1.3in]{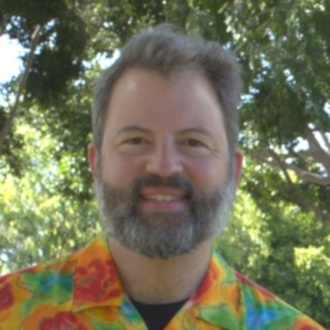}&
\includegraphics[width=1.3in]{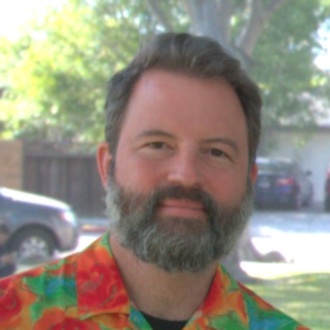} &
\includegraphics[width=1.3in]{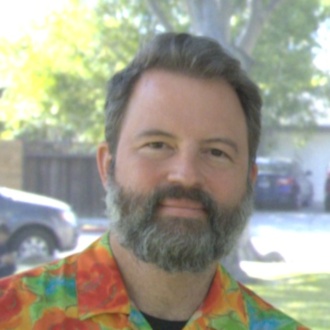} &
\includegraphics[width=1.3in]{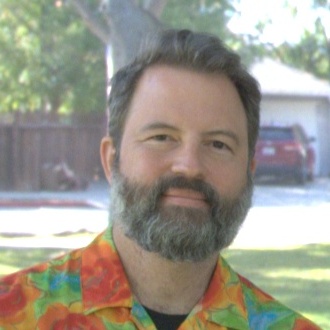} &
\includegraphics[width=1.3in]{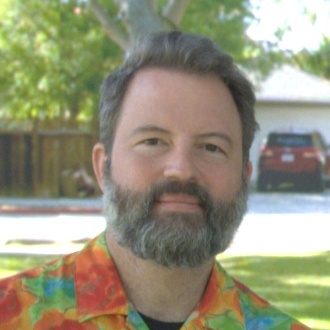}\\

{\rotatebox{90}{\footnotesize{lit chart}}} &
\includegraphics[width=1.3in]{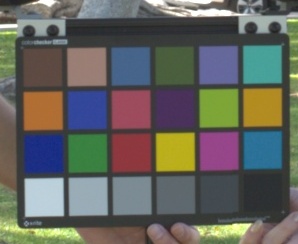}&
\includegraphics[width=1.3in]{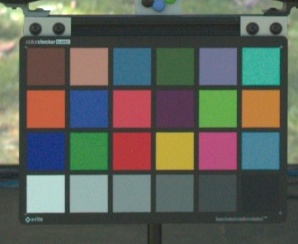} &
\includegraphics[width=1.3in]{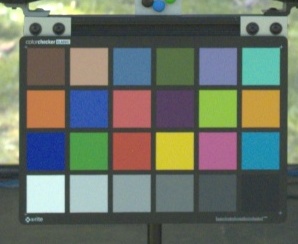} &
\includegraphics[width=1.3in]{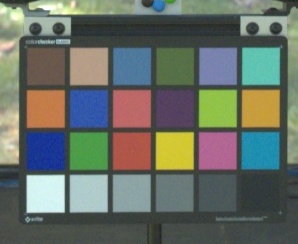} &
\includegraphics[width=1.3in]{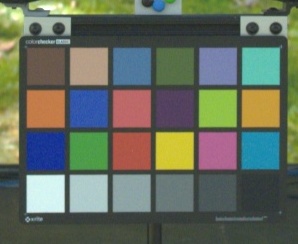}\\

 &
{\rotatebox{90}{\footnotesize{lit (compared)}}} &
\includegraphics[width=1.3in]{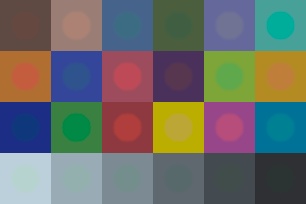} &
\includegraphics[width=1.3in]{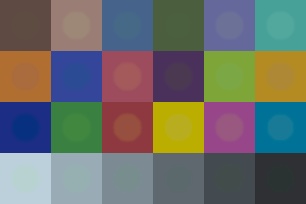} &
\includegraphics[width=1.3in]{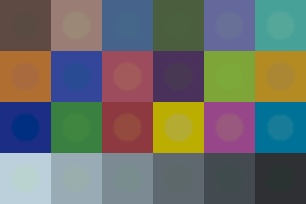} &
\includegraphics[width=1.3in]{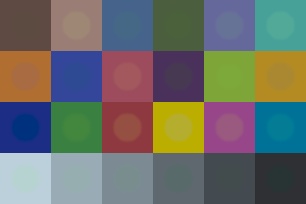}\\

 &
{\rotatebox{90}{\footnotesize{displayed chart}}} &
\includegraphics[width=1.3in]{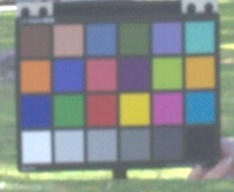} &
\includegraphics[width=1.3in]{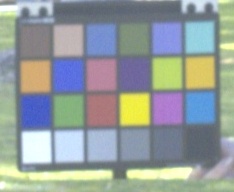} &
\includegraphics[width=1.3in]{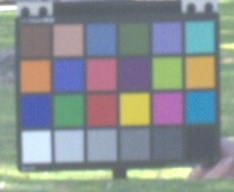} &
\includegraphics[width=1.3in]{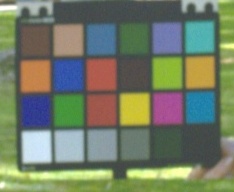}\\

 &
{\rotatebox{90}{\footnotesize{displayed (compared)}}} &
\includegraphics[width=1.3in]{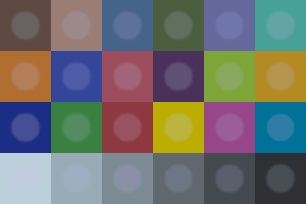} &
\includegraphics[width=1.3in]{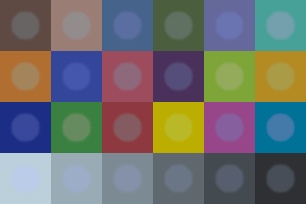} &
\includegraphics[width=1.3in]{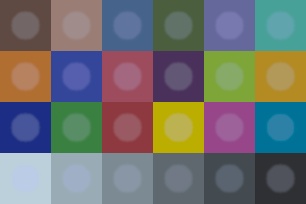} &
\includegraphics[width=1.3in]{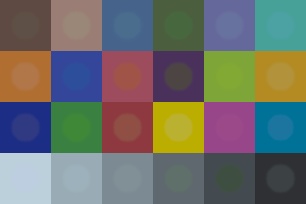}\\

& 
\multicolumn{1}{c}{\footnotesize{(a) real environment}} & 
\footnotesize{(b) VP: $\mathbf{M}$ only baseline} & 
\footnotesize{(c) VP: $\mathbf{M}$ and $\mathbf{Q}$} &
\footnotesize{(d) VP: $\mathbf{M}$, $\mathbf{N}$, and $\mathbf{Q}$} &
\footnotesize{(e) VP: $\mathbf{M}$, $\mathbf{N}$, $\mathbf{Q}$, and black level} \\

& 
\multicolumn{1}{c}{\footnotesize{(outdoors in shade)}} &
\footnotesize{(previous approaches)} &
\footnotesize{(no $\mathbf{Q}^{-1}$ correction)} &
\footnotesize{(all matrices)} &
\footnotesize{(our full approach)} \\

\end{tabular}
\vspace{-5pt}
\caption{Column (a): two subjects photographed outdoors [Fig. \ref{fig:panos}(6)] and select crops. Column (b): VP lighting reproduction using the baseline approach ($\mathbf{M}$ only). Column (c): applying post-correction $\mathbf{Q}$ to the images of column (b). Column (d): also pre-correcting in-frustum content with $\mathbf{N}$. Column (e): including our black level subtraction. For \textit{comparison} charts, background squares are sampled from the chart in the real environment, while foreground circles are sampled from corresponding charts in each column. In the VP stage we photographed both a chart \textit{lit} by out-of-frustum content and \textit{displayed} on the panels. Compared with the baseline (b), $\mathbf{Q}$ enables improved color rendition for the \textit{lit} chart, desaturating skin tones and improving the appearance of orange/yellow materials. Black level subtraction provides a dramatic improvement to \textit{displayed} content.}
\vspace{-5pt}
\label{fig:fullprocess}
\end{figure*}

%% file: all_results_figure1.tex
\begin{figure*}[p]
\vspace{0pt}
\begin{tabular}{@{}c@{}c@{}c@{}c@{}c@{}}

\multicolumn{1}{@{}c@{}}{\includegraphics[width=2.3in]{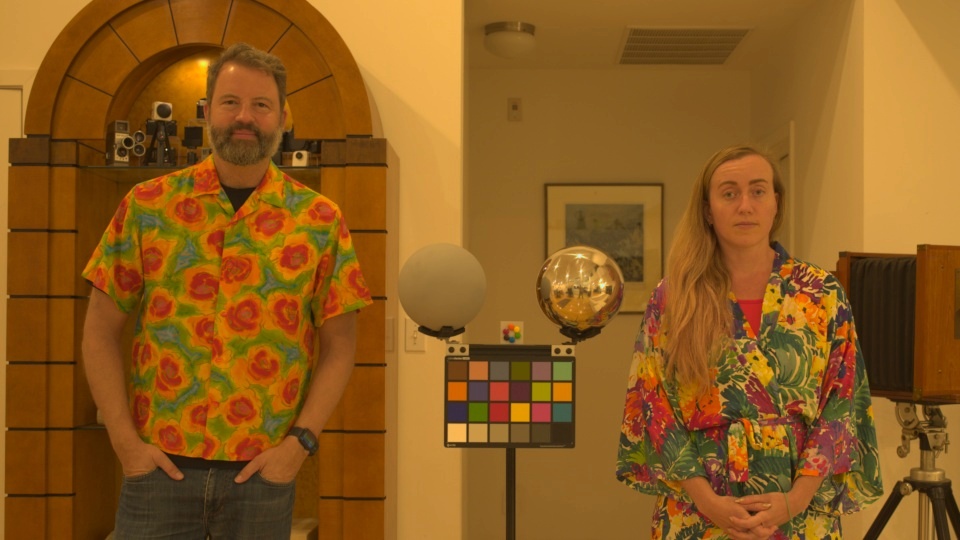}} &
\multicolumn{2}{@{ }c@{ }}{\includegraphics[width=2.3in]{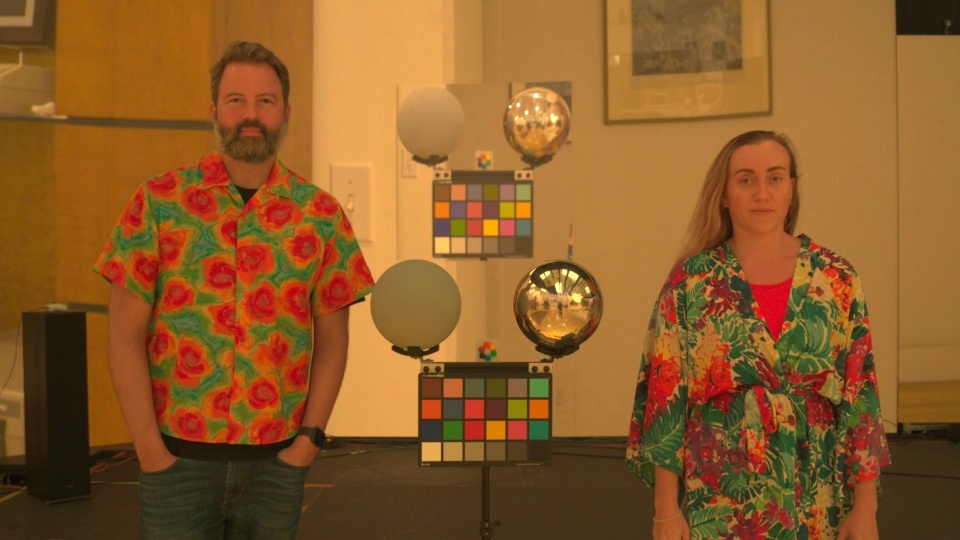}} &
\multicolumn{2}{@{}c@{}}{\includegraphics[width=2.3in]{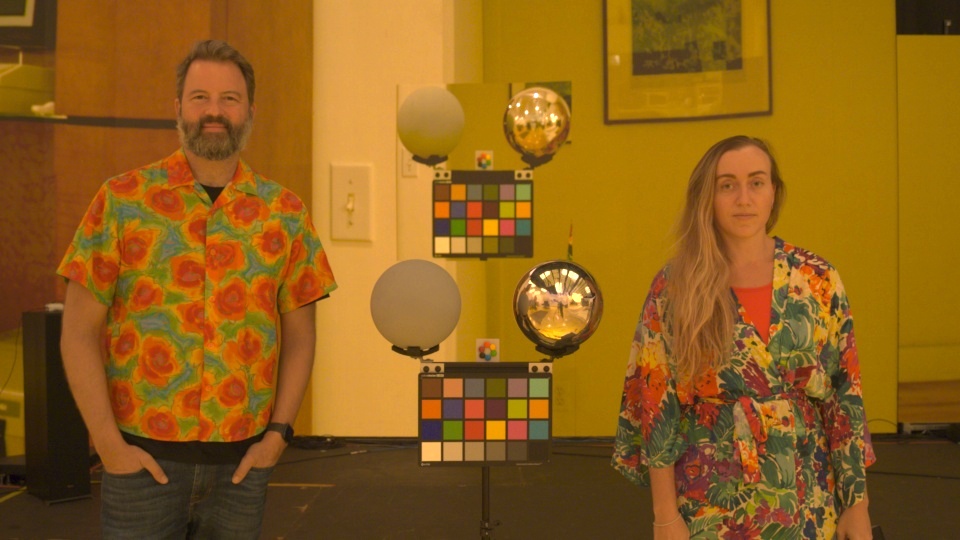}} \\

\multicolumn{1}{@{}c@{}}{\includegraphics[height=0.75in]{images/panos/houselights_kitchen.jpg}} &
\multicolumn{1}{@{ }c@{}}{\includegraphics[width=1.135in]{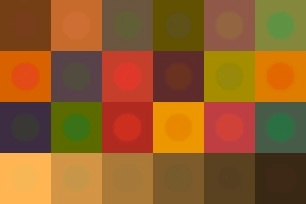}} &
\multicolumn{1}{@{}c@{}}{\includegraphics[width=1.135in]{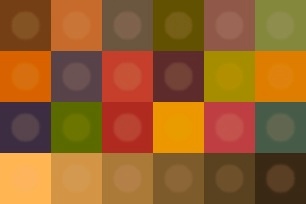}} &
\multicolumn{1}{@{}c@{}}{\includegraphics[width=1.135in]{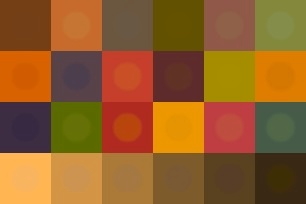}} &
\multicolumn{1}{@{ }c@{}}{\includegraphics[width=1.135in]{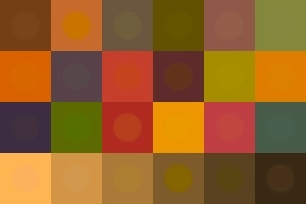}} \\

\multicolumn{1}{@{}c@{}}{\footnotesize{HDR IBL (warm white LED)}} &
\multicolumn{1}{@{ }c@{}}{\footnotesize{lit comparison}} &
\multicolumn{1}{@{}c@{}}{\footnotesize{displayed comparison}} &
\multicolumn{1}{@{}c@{}}{\footnotesize{lit comparison}} &
\multicolumn{1}{@{ }c@{}}{\footnotesize{displayed comparison}} \\

& & & & \\

\multicolumn{1}{@{}c@{}}{\includegraphics[width=2.3in]{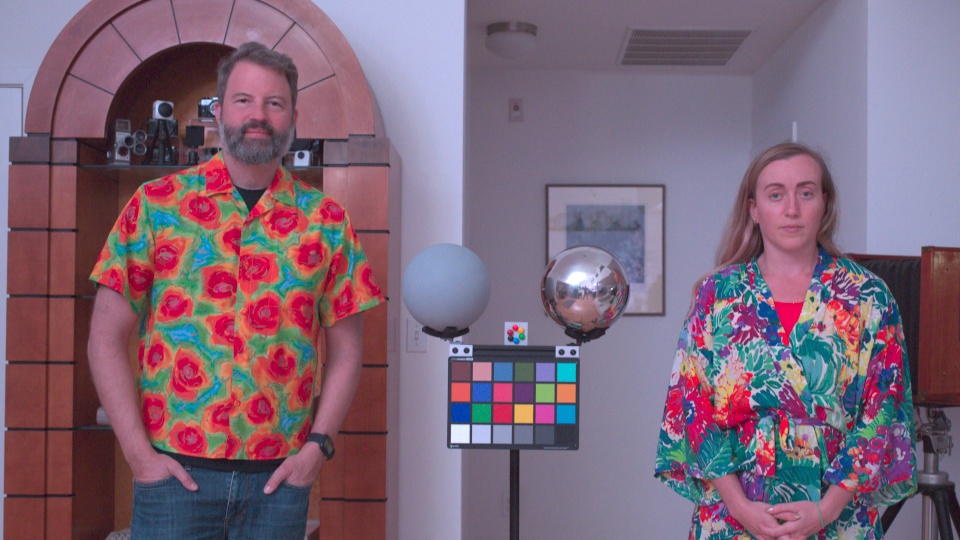}} &
\multicolumn{2}{@{ }c@{ }}{\includegraphics[width=2.3in]{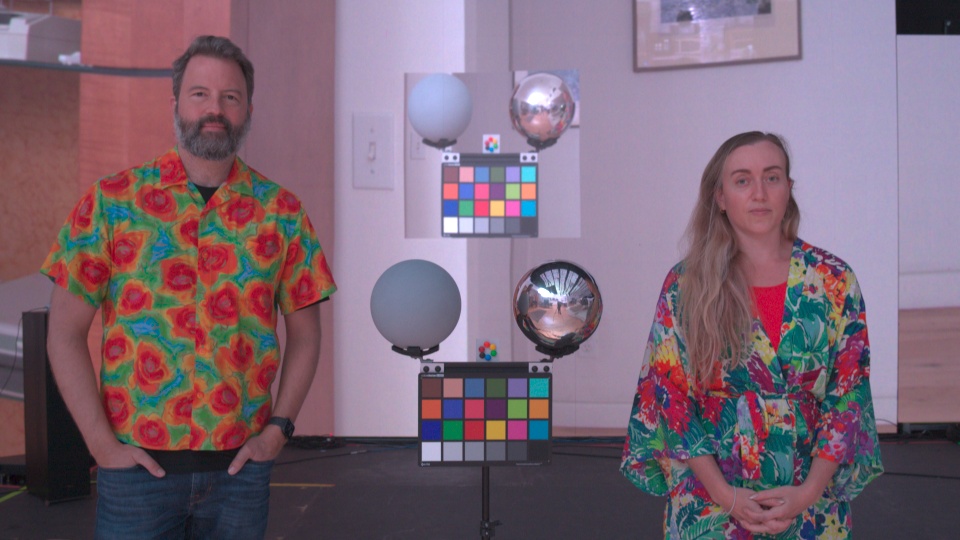}} &
\multicolumn{2}{@{}c@{}}{\includegraphics[width=2.3in]{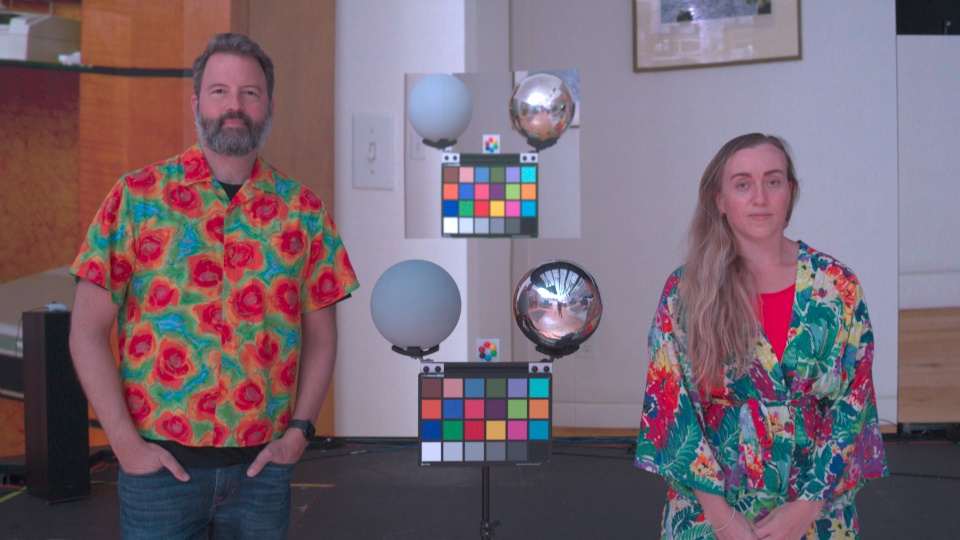}} \\

\multicolumn{1}{@{}c@{}}{\includegraphics[height=0.75in]{images/panos/rgbled_kitchen.jpg}} &
\multicolumn{1}{@{ }c@{}}{\includegraphics[width=1.135in]{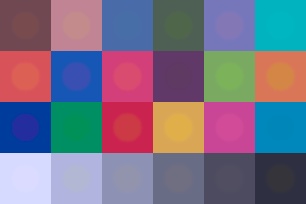}} &
\multicolumn{1}{@{}c@{}}{\includegraphics[width=1.135in]{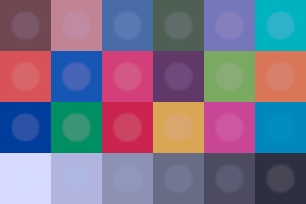}} &
\multicolumn{1}{@{}c@{}}{\includegraphics[width=1.135in]{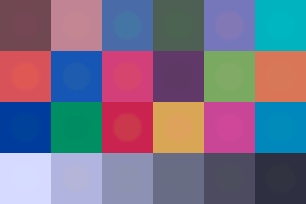}} &
\multicolumn{1}{@{ }c@{}}{\includegraphics[width=1.135in]{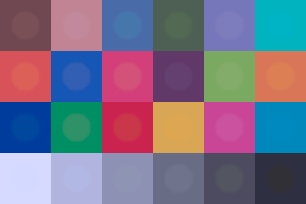}} \\

\multicolumn{1}{@{}c@{}}{\footnotesize{HDR IBL (RGB LED based white)}} &
\multicolumn{1}{@{ }c@{}}{\footnotesize{lit comparison}} &
\multicolumn{1}{@{}c@{}}{\footnotesize{displayed comparison}} &
\multicolumn{1}{@{}c@{}}{\footnotesize{lit comparison}} &
\multicolumn{1}{@{ }c@{}}{\footnotesize{displayed comparison}} \\

& & & & \\

\multicolumn{1}{@{}c@{}}{\includegraphics[width=2.3in]{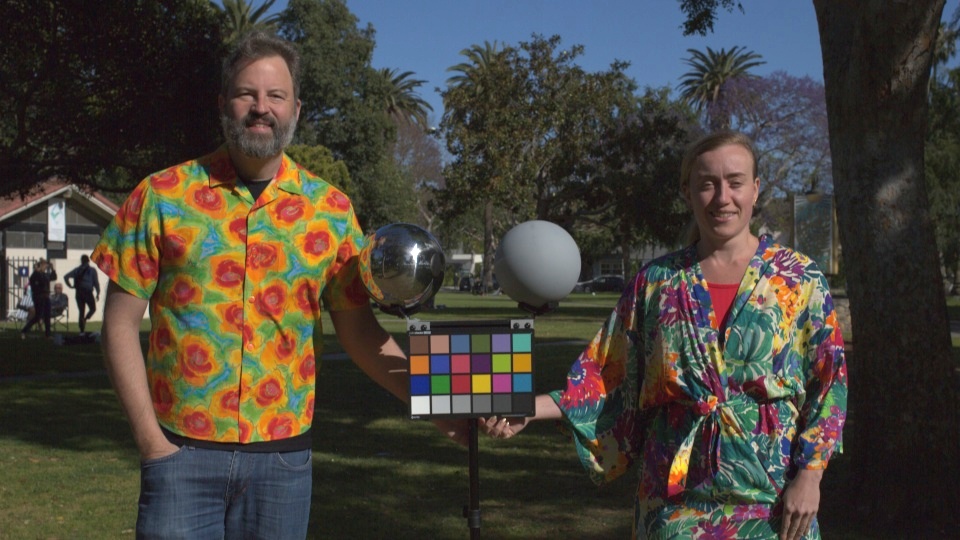}} &
\multicolumn{2}{@{ }c@{ }}{\includegraphics[width=2.3in]{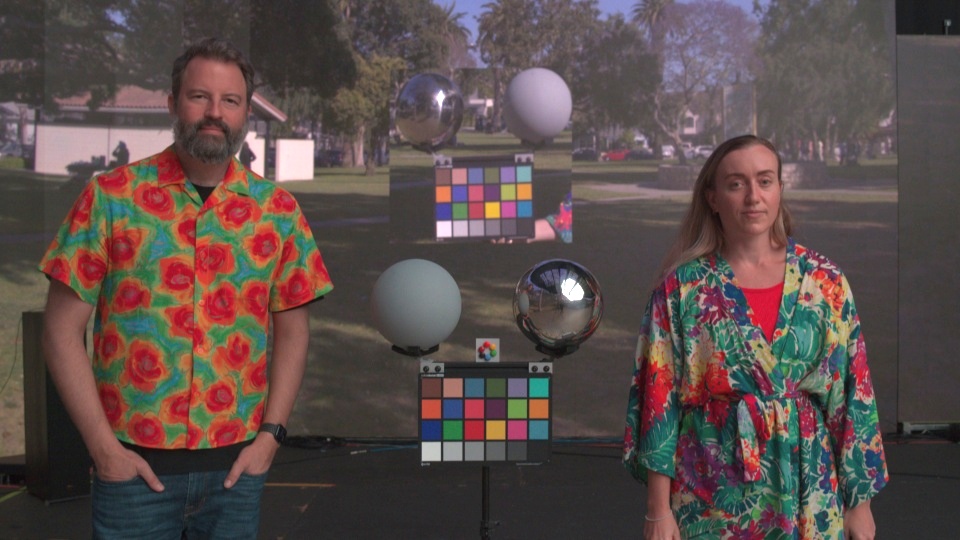}} &
\multicolumn{2}{@{}c@{}}{\includegraphics[width=2.3in]{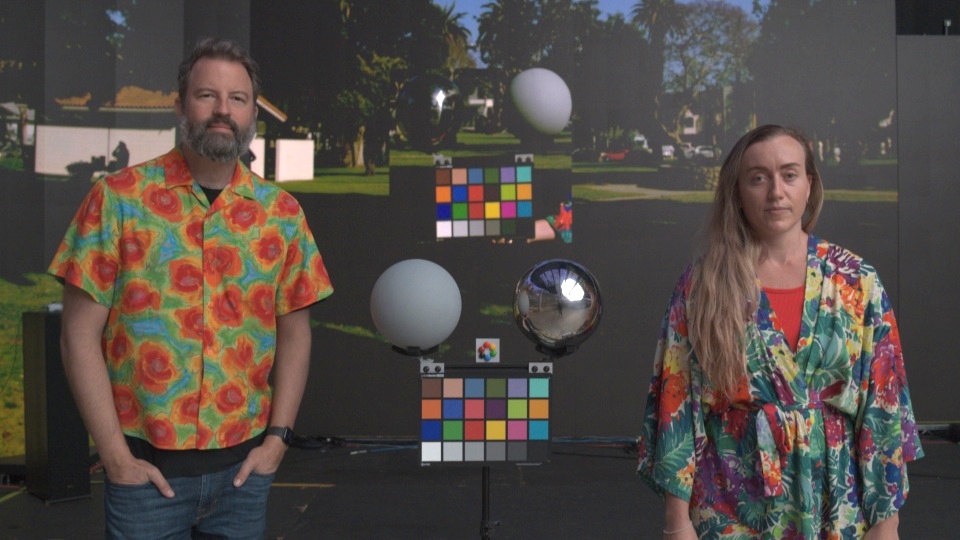}} \\

\multicolumn{1}{@{}c@{}}{\includegraphics[height=0.75in]{images/panos/sunny_park.jpg}} &
\multicolumn{1}{@{ }c@{}}{\includegraphics[width=1.135in]{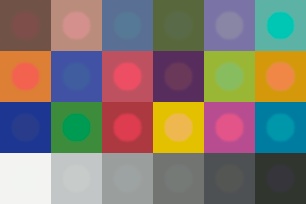}} &
\multicolumn{1}{@{}c@{}}{\includegraphics[width=1.135in]{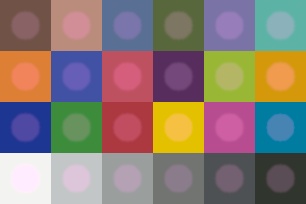}} &
\multicolumn{1}{@{}c@{}}{\includegraphics[width=1.135in]{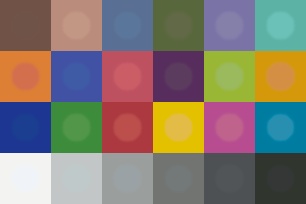}} &
\multicolumn{1}{@{ }c@{}}{\includegraphics[width=1.135in]{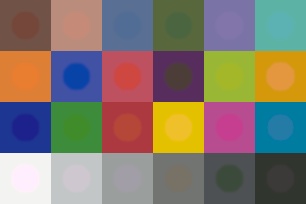}} \\

\multicolumn{1}{@{}c@{}}{\footnotesize{HDR IBL (outdoors, direct sunlight)}} &
\multicolumn{1}{@{ }c@{}}{\footnotesize{lit comparison}} &
\multicolumn{1}{@{}c@{}}{\footnotesize{displayed comparison}} &
\multicolumn{1}{@{}c@{}}{\footnotesize{lit comparison}} &
\multicolumn{1}{@{ }c@{}}{\footnotesize{displayed comparison}} \\

& & & & \\
\multicolumn{1}{@{}c@{}}{\footnotesize{(a) photograph in real environment}} &
\multicolumn{2}{@{ }c@{}}{\footnotesize{(b) VP using primary-based calibration, $\textbf{M}$ only (baseline)}} &
\multicolumn{2}{@{}c@{}}{\footnotesize{(c) VP using $\textbf{M}$, $\textbf{N}$, $\textbf{Q}$, and black level (our approach)}} \\

\end{tabular}
\vspace{-5pt}
\caption{For three spectrally-diverse lighting environments, we show subjects photographed in the real world (a), lighting reproduction in a VP stage using the baseline approach (b), and using our full approach (c). Compared with the baseline calibration method, our approach enables improved color rendition for the \textit{lit} chart, desaturating skin tones and improving the appearance of orange/yellow materials. Our black level subtraction removes the appearance of light bounced off the in-camera background LED panels. Observe the color rendition for the RGB LED based lighting environment (middle rows) is already quite good as we are asking RGB LED panels to reproduce RGB LED based illumination.}
\vspace{-5pt}
\label{fig:all_env_results1}
\end{figure*}

%% file: all_results_figure2.tex
\begin{figure*}[p]
\vspace{0pt}
\begin{tabular}{@{}c@{}c@{}c@{}c@{}c@{}}

\multicolumn{1}{@{}c@{}}{\includegraphics[width=2.3in]{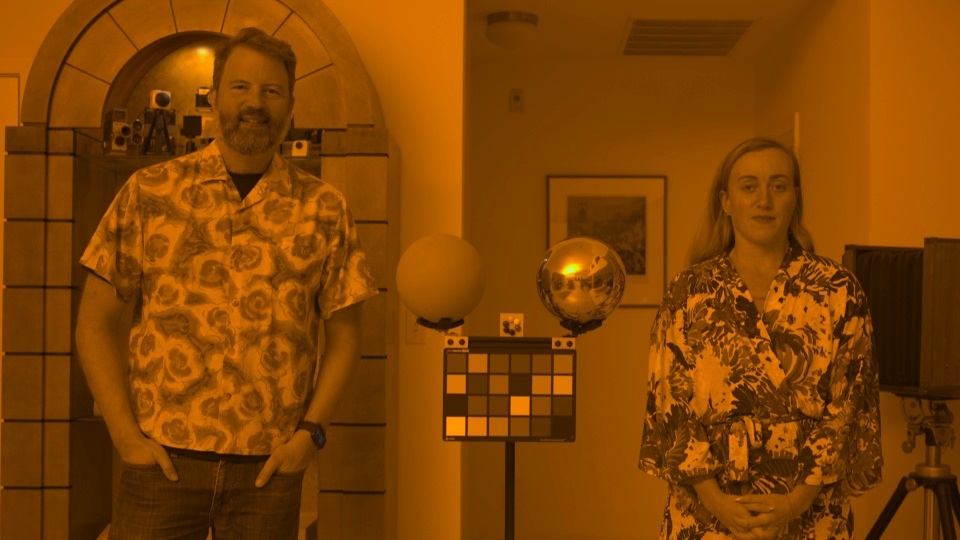}} &
\multicolumn{2}{@{ }c@{ }}{\includegraphics[width=2.3in]{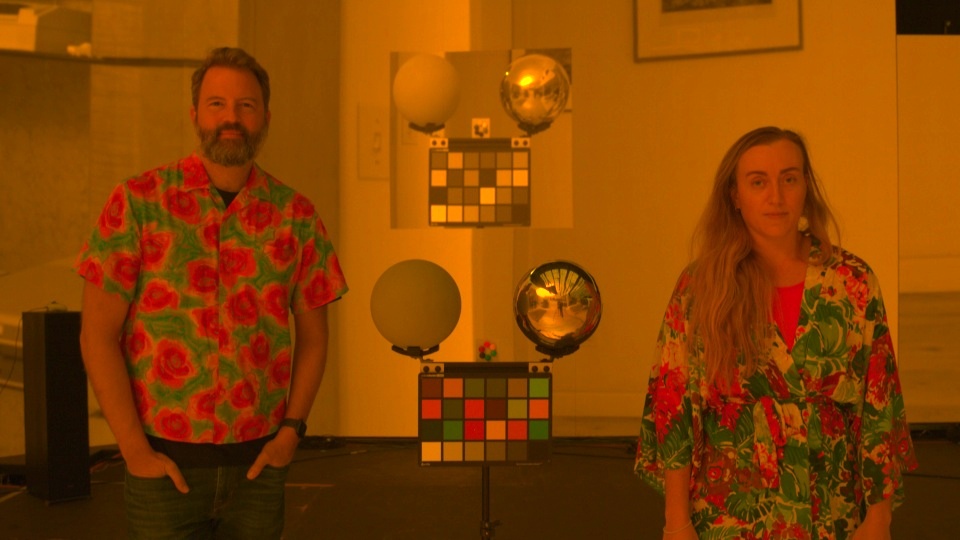}} &
\multicolumn{2}{@{}c@{}}{\includegraphics[width=2.3in]{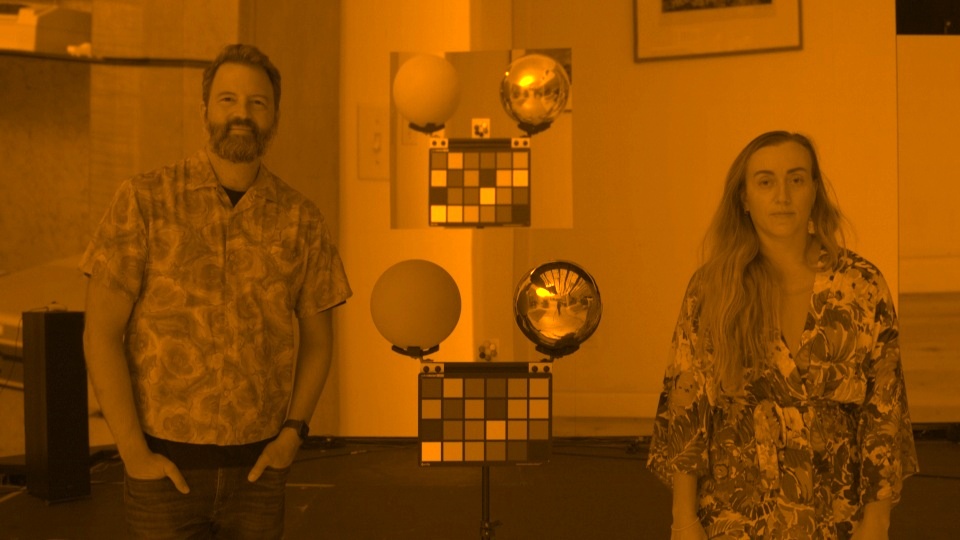}} \\

\multicolumn{1}{@{}c@{}}{\includegraphics[height=0.75in]{images/panos/sodium_vapor.jpg}} &
\multicolumn{1}{@{ }c@{}}{\includegraphics[width=1.135in]{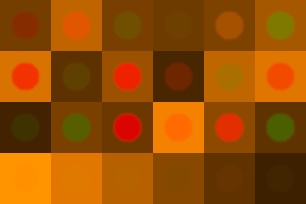}} &
\multicolumn{1}{@{}c@{}}{\includegraphics[width=1.135in]{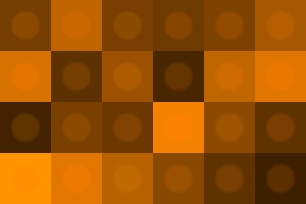}} &
\multicolumn{1}{@{}c@{}}{\includegraphics[width=1.135in]{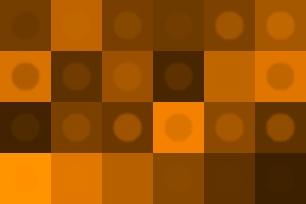}} &
\multicolumn{1}{@{ }c@{}}{\includegraphics[width=1.135in]{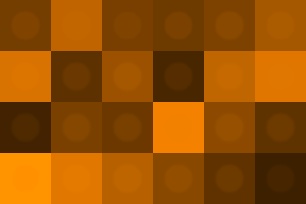}} \\

\multicolumn{1}{@{}c@{}}{\footnotesize{HDR IBL (sodium vapor)}} &
\multicolumn{1}{@{ }c@{}}{\footnotesize{lit comparison}} &
\multicolumn{1}{@{}c@{}}{\footnotesize{displayed comparison}} &
\multicolumn{1}{@{}c@{}}{\footnotesize{lit comparison}} &
\multicolumn{1}{@{ }c@{}}{\footnotesize{displayed comparison}} \\

& & & & \\

\multicolumn{1}{@{}c@{}}{\includegraphics[width=2.3in]{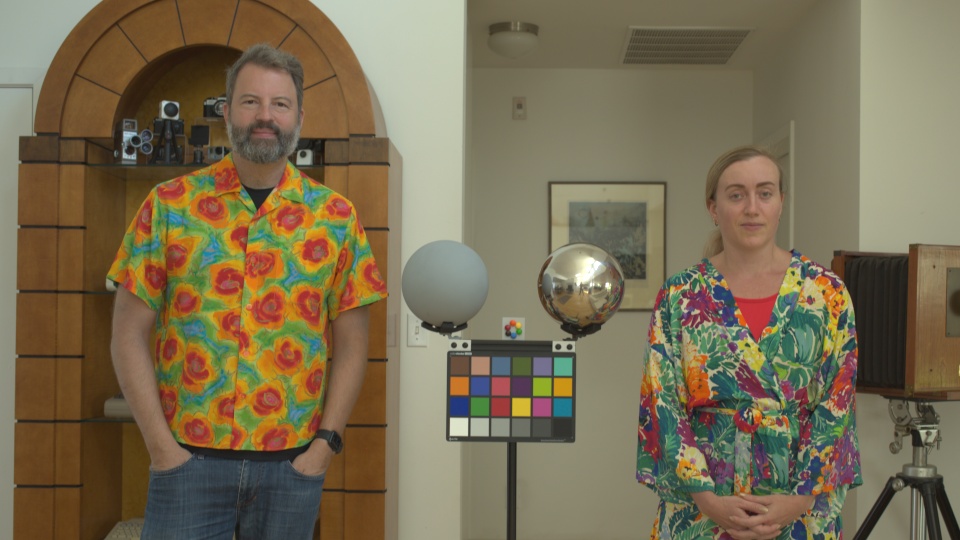}} &
\multicolumn{2}{@{ }c@{ }}{\includegraphics[width=2.3in]{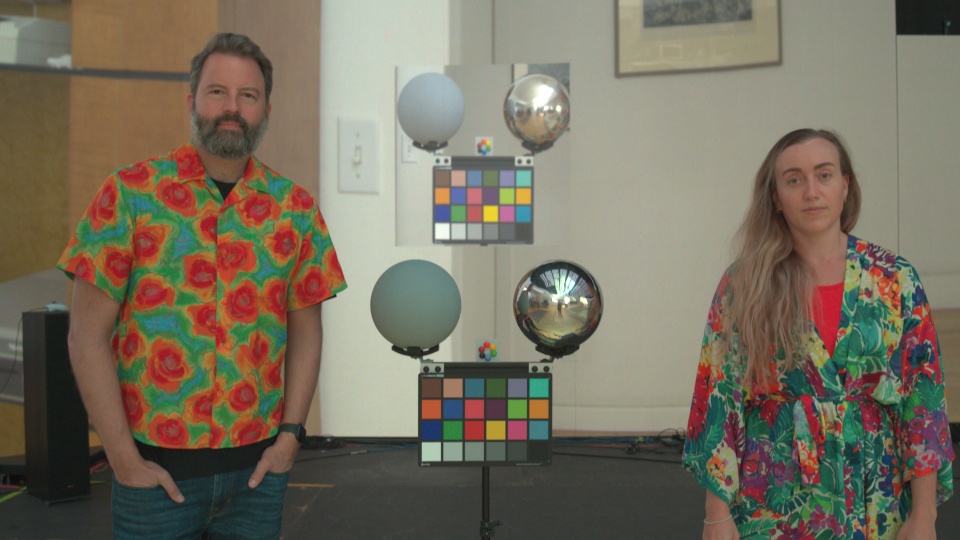}} &
\multicolumn{2}{@{}c@{}}{\includegraphics[width=2.3in]{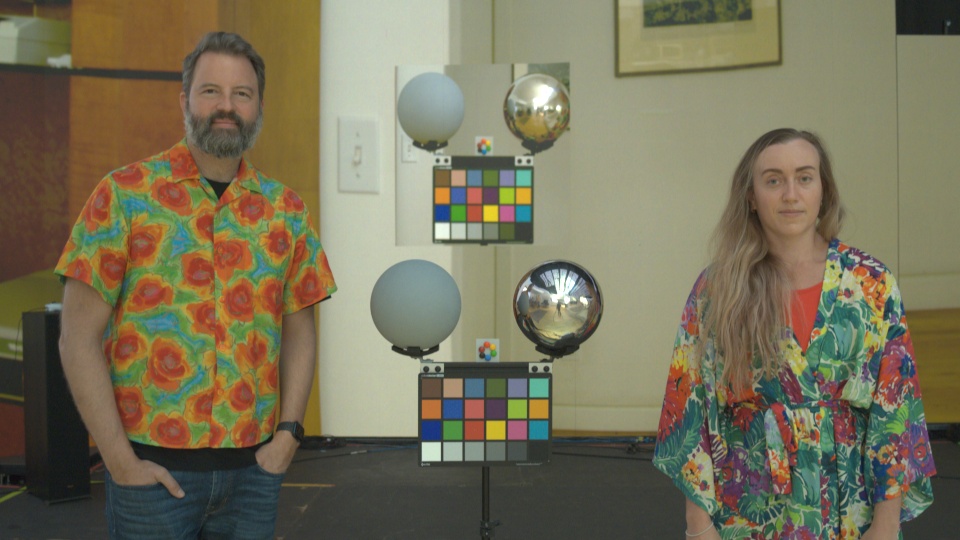}} \\

\multicolumn{1}{@{}c@{}}{\includegraphics[height=0.75in]{images/panos/naturallights_kitchen.jpg}} &
\multicolumn{1}{@{ }c@{}}{\includegraphics[width=1.135in]{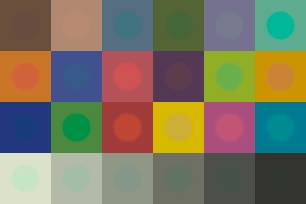}} &
\multicolumn{1}{@{}c@{}}{\includegraphics[width=1.135in]{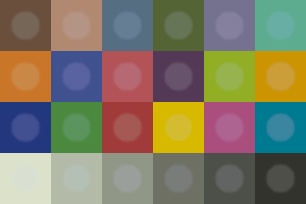}} &
\multicolumn{1}{@{}c@{}}{\includegraphics[width=1.135in]{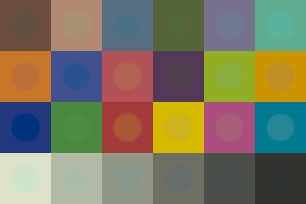}} &
\multicolumn{1}{@{ }c@{}}{\includegraphics[width=1.135in]{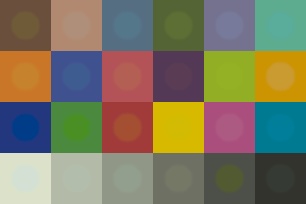}} \\

\multicolumn{1}{@{}c@{}}{\footnotesize{HDR IBL (natural daylight)}} &
\multicolumn{1}{@{ }c@{}}{\footnotesize{lit comparison}} &
\multicolumn{1}{@{}c@{}}{\footnotesize{displayed comparison}} &
\multicolumn{1}{@{}c@{}}{\footnotesize{lit comparison}} &
\multicolumn{1}{@{ }c@{}}{\footnotesize{displayed comparison}} \\

& & & & \\

\multicolumn{1}{@{}c@{}}{\includegraphics[width=2.3in]{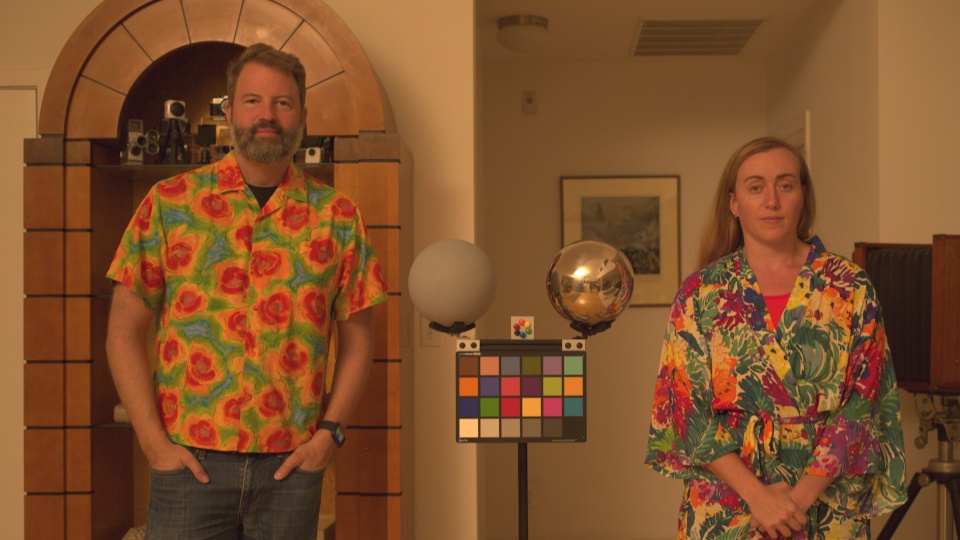}} &
\multicolumn{2}{@{ }c@{ }}{\includegraphics[width=2.3in]{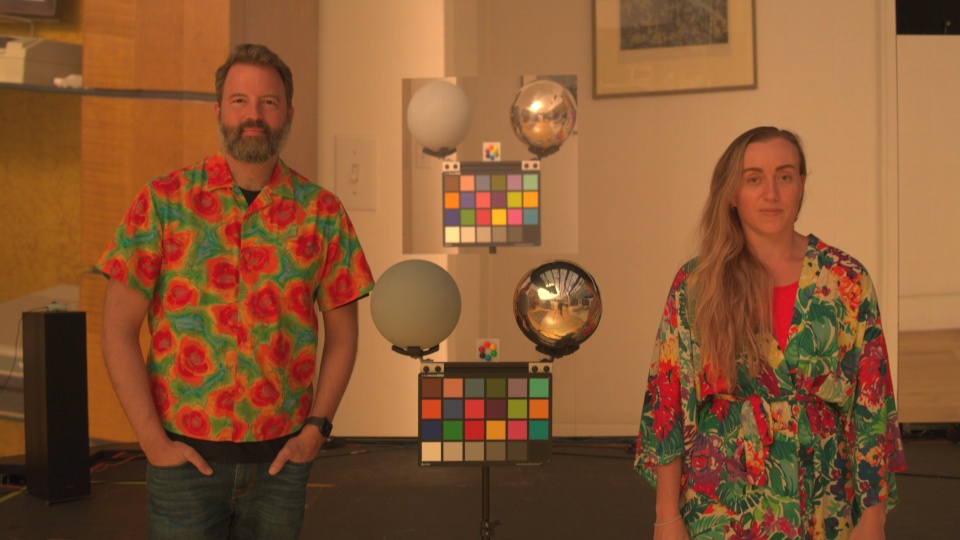}} &
\multicolumn{2}{@{}c@{}}{\includegraphics[width=2.3in]{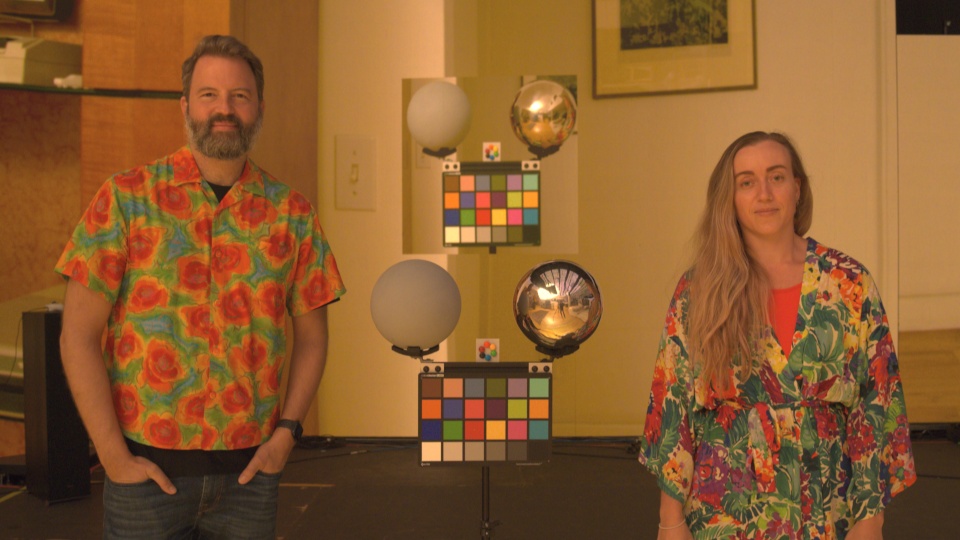}} \\

\multicolumn{1}{@{}c@{}}{\includegraphics[height=0.75in]{images/panos/incandescent_kitchen.jpg}} &
\multicolumn{1}{@{ }c@{}}{\includegraphics[width=1.135in]{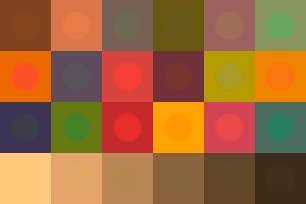}} &
\multicolumn{1}{@{}c@{}}{\includegraphics[width=1.135in]{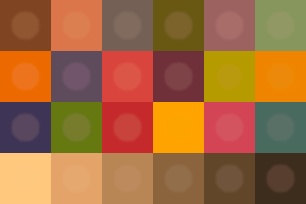}} &
\multicolumn{1}{@{}c@{}}{\includegraphics[width=1.135in]{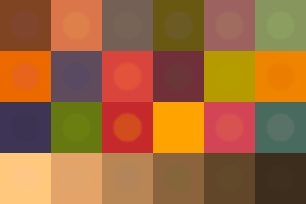}} &
\multicolumn{1}{@{ }c@{}}{\includegraphics[width=1.135in]{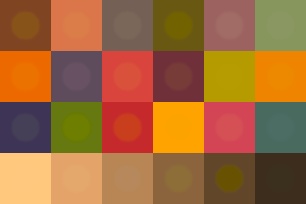}} \\

\multicolumn{1}{@{}c@{}}{\footnotesize{HDR IBL (incandescent/tungsten)}} &
\multicolumn{1}{@{ }c@{}}{\footnotesize{lit comparison}} &
\multicolumn{1}{@{}c@{}}{\footnotesize{displayed comparison}} &
\multicolumn{1}{@{}c@{}}{\footnotesize{lit comparison}} &
\multicolumn{1}{@{ }c@{}}{\footnotesize{displayed comparison}} \\

& & & & \\
\multicolumn{1}{@{}c@{}}{\footnotesize{(a) photograph in real environment}} &
\multicolumn{2}{@{ }c@{}}{\footnotesize{(b) VP using primary-based calibration, $\textbf{M}$ only (baseline)}} &
\multicolumn{2}{@{}c@{}}{\footnotesize{(c) VP using $\textbf{M}$, $\textbf{N}$, $\textbf{Q}$, and black level (our approach)}} \\

\end{tabular}
\vspace{-5pt}
\caption{For three spectrally-diverse lighting environments, we show subjects photographed in the real world (a), lighting reproduction in a VP stage using the baseline approach (b), and using our full approach (c). Compared with the baseline calibration method, our approach enables improved color rendition for the \textit{lit} chart, desaturating skin tones and improving the appearance of orange/yellow materials. Our black level subtraction removes the appearance of light bounced off the in-camera background LED panels. For the sodium vapor lighting environment of the top row, $\textbf{Q}$ is able to completely desaturate the RGB-LED color chart in the VP stage to better match the appearance of the nearly monochromatic sodium vapor illumination in the real world.}
\vspace{-5pt}
\label{fig:all_env_results2}
\end{figure*}